\documentclass[journal ]{new-aiaa}
\usepackage[utf8]{inputenc}
\usepackage{textcomp}

\usepackage{graphicx}
\usepackage{amsmath}
\usepackage[version=4]{mhchem}
\usepackage{siunitx}
\usepackage{longtable,tabularx}
\usepackage{subcaption}
\usepackage{float}
\usepackage{algorithm}
\usepackage{algpseudocode}

\usepackage{amssymb}
\usepackage{bm}
\usepackage{hyperref}

\newcommand{\tp}{\mathrm{T}}

\title{AI-Augmented Model Predictive Control for Safe and Adaptive Rendezvous and Proximity Operations}

\author{
Luca Sportelli\footnote{Ph.D. Student, Department of Aerospace Engineering, Embry--Riddle Aeronautical University, Daytona Beach, FL 32114. Corresponding author: sportell@my.erau.edu},
Tyler Barr\footnote{Undergraduate Student, Department of Aerospace Engineering, Embry--Riddle Aeronautical University, Daytona Beach, FL 32114.},
Cagri Kilic\footnote{Assistant Professor, Department of Aerospace Engineering, Embry--Riddle Aeronautical University, Daytona Beach, FL 32114.},
and Di Wu\footnote{Assistant Professor, Department of Aerospace Engineering, Embry--Riddle Aeronautical University, Daytona Beach, FL 32114.}
}

\affil{Department of Aerospace Engineering, Embry-Riddle Aeronautical University, Daytona Beach, FL 32114}

\begin{document}

\maketitle

\begin{abstract}
Autonomous rendezvous and proximity operations (RPO) in adversarial orbital environments require guidance architectures balancing target pursuit, safety preservation, and real-time adaptability under dynamically evolving interaction conditions. Although learning-based approaches show promise, their application to safety-critical orbital robotics remains limited by concerns regarding interpretability, robustness, and constraint awareness. This work presents an adaptive Model Predictive Control (MPC) framework for autonomous spacecraft RPO in multi-agent adversarial scenarios. The proposed architecture combines a constrained receding-horizon MPC formulation with a data-driven supervisory tuning layer that adjusts controller parameters from offline closed-loop evaluation and online interaction geometry. Relative motion follows Clohessy–Wiltshire (CW) dynamics, enabling computationally efficient finite-horizon prediction and real-time quadratic optimization. The MPC formulation incorporates actuator limits, predictive keep-out-zone constraints, slack-variable feasibility handling, and optional Control Barrier Function (CBF) safety filtering. Rather than generating thrust commands directly, the adaptive layer modifies interpretable MPC parameters, including tracking weights, safety penalties, minimum-separation objectives, and keep-out-zone objectives. The framework was evaluated in the official Kerbal Space Program Differential Game (KSPDG) Capture-the-Satellite environment through Monte Carlo simulations. Results demonstrate improved closed-loop robustness, adaptive maneuvering behavior, and rendezvous performance compared with fixed-parameter MPC while preserving safety-aware operation and real-time feasibility, providing a modular, interpretable foundation for adaptive spacecraft RPO.

\end{abstract}

\section*{Nomenclature}

{\renewcommand\arraystretch{1.0}
\begin{longtable*}{@{}l @{\quad=\quad} l@{}}

$A_c$ & continuous-time state matrix \\
$A_d$ & discrete-time state transition matrix \\
$B_c$ & continuous-time control matrix \\
$B_d$ & discrete-time control matrix \\
$d_{BG}$ & Bandit--Guard separation distance, m \\
$d_{BL}$ & Bandit--Lady separation distance, m \\
$d_{\mathrm{KOZ}}$ & keep-out-zone radius, m \\
$d_{\min}$ & minimum-separation objective, m \\
$d_{\min,\mathrm{eff}}$ & effective minimum-separation objective, m \\
$H_p$ & prediction horizon \\
$H_u$ & control horizon \\
$K_{\mathrm{KOZ}}$ & active KOZ prediction horizon \\
$h$ & Control Barrier Function safety function \\
$J$ & MPC objective function \\
$k_0,k_1$ & Control Barrier Function gains \\
$n$ & orbital mean motion, rad/s \\
$P_k$ & supervisory parameter vector \\
$P^\star$ & offline-optimized supervisory parameter vector \\
$Q$ & state-tracking weighting matrix \\
$R$ & control-effort weighting matrix \\
$r_0$ & reference-orbit radius, m \\
$s_i$ & slack variable \\
$T$ & control-horizon lifting matrix \\
$u_{\max}$ & actuator acceleration limit \\
$\mathbf{u}$ & control input vector \\
$\mathbf{U}_k$ & stacked control decision sequence \\
$w_U$ & control-effort weight \\
$w_{\mathrm{KOZ}}$ & KOZ safety weight \\
$w_{L,\mathrm{pos}}$ & Lady position-tracking weight \\
$w_{L,\mathrm{vel}}$ & Lady velocity-tracking weight \\
$\mathbf{x}$ & spacecraft state vector \\
$\mathbf{X}_k$ & stacked predicted state sequence \\
$\mathbf{z}$ & QP decision vector \\
$\Phi$ & stacked state-transition prediction matrix \\
$\Gamma$ & stacked control prediction matrix \\
$\gamma_K$ & KOZ constraint decay coefficient \\
$\Delta t$ & controller update timestep, s \\
$\lambda_s$ & slack-variable penalty weight \\
$\mu$ & gravitational parameter, m$^3$/s$^2$ \\
$\sigma_{H_p}$ & terminal-cost multiplier \\

\multicolumn{2}{@{}l}{\textbf{Subscripts}} \\

$B$ & Bandit spacecraft \\
$G$ & Guard spacecraft \\
$L$ & Lady spacecraft \\
KOZ & keep-out zone \\

\end{longtable*}}

\addtocounter{table}{-1}

\section{Introduction}

Autonomous RPO remain a central challenge in orbital robotics, particularly in adversarial or uncooperative scenarios where safety, performance, and adaptability must be balanced in real time. These operations are becoming increasingly relevant for applications including on-orbit servicing, inspection, debris mitigation, and infrastructure protection in congested orbital environments.

MPC has been widely adopted in robotics and aerospace systems due to its ability to explicitly enforce constraints while optimizing future system behavior over a receding horizon \cite{rawlings2017mpc,MAYNE2000789}. In orbital robotics, MPC formulations based on CW or Hill dynamics provide computationally efficient guidance strategies compatible with real-time operation and constrained maneuvering \cite{camacho2007mpc,pesce_modern_2022,starek2017fast}.

At the same time, learning-based approaches have demonstrated promising capabilities in adaptive guidance, trajectory optimization, and multi-agent interaction problems. Recent studies have highlighted the growing role of artificial intelligence techniques within spacecraft guidance, navigation, and control architectures, particularly for autonomous decision-making and adaptive mission operations \cite{izzo2018surveyartificialintelligencetrends}. However, the direct deployment of end-to-end learning architectures in safety-critical orbital environments remains challenging due to limited interpretability, robustness concerns, and the difficulty of guaranteeing constraint-aware behavior.

This work proposes an adaptive MPC framework for autonomous RPO in adversarial multi-agent scenarios. The proposed architecture preserves a constrained receding-horizon MPC structure throughout execution, while an outer-loop adaptive layer modifies selected optimization-level parameters governing the predictive controller behavior. The proposed architecture combines linearized orbital dynamics, constrained convex optimization, and adaptive parameter tuning into a modular and interpretable control framework.

\begin{figure}[hbt!]
\centering
\includegraphics[width=\textwidth]{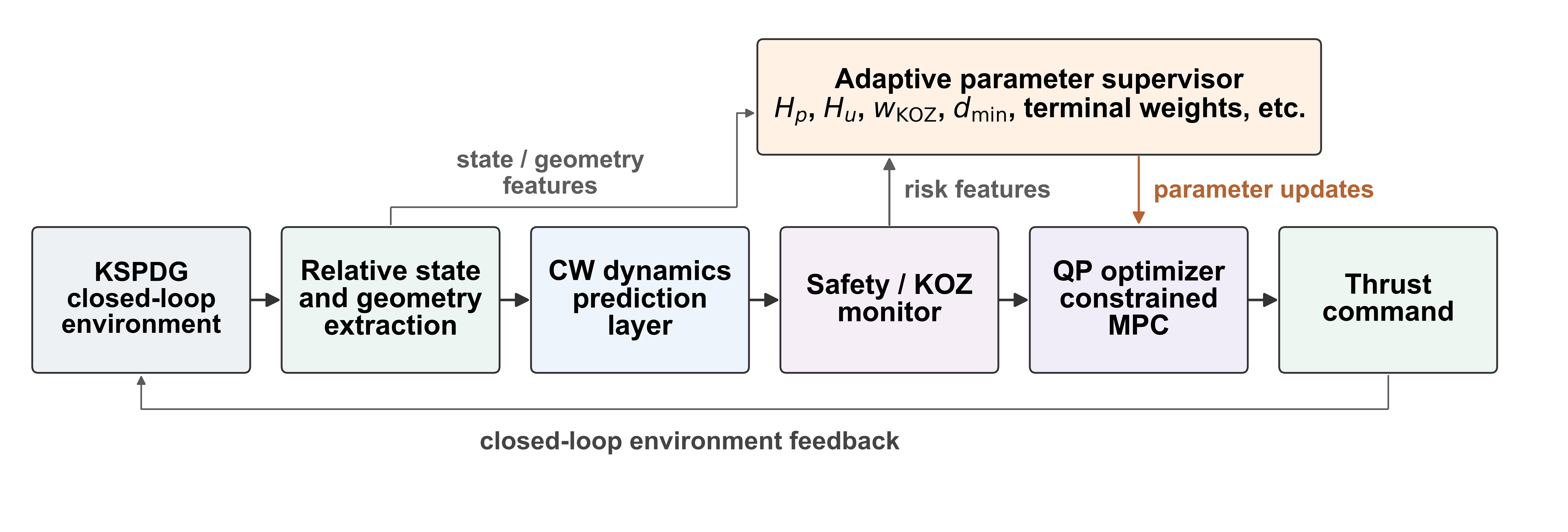}
\caption{Overview of the proposed AI-augmented MPC framework for adversarial RPO.}
\label{fig:architectureMPC}
\end{figure}

Figure~\ref{fig:architectureMPC} provides a visual overview of the proposed AI-augmented MPC framework, highlighting the interaction between the constrained receding-horizon MPC layer, predictive safety mechanisms, and the adaptive supervisory parameter-update process. State and geometry features extracted from the orbital interaction are processed by the adaptive supervisory layer, which updates interpretable MPC optimization parameters using safety and KOZ monitoring information. The constrained MPC/QP layer remains responsible for generating spacecraft thrust commands under orbital dynamics, actuator, and predictive safety constraints.

The framework is organized as two clearly separated loops. An \emph{inner loop} solves a constrained receding-horizon quadratic program at every controller tick using the current measured state, fixed dynamics model, and active constraint set (Eq.~\ref{eq:parametric_qp}). An \emph{outer loop} updates the parameter vector $P_k$ of the inner-loop optimization problem through offline parameter optimization and online supervisory adaptation based on the evolving orbital interaction geometry and safety conditions. The inner-loop dynamics model, actuator bounds, and predictive safety constraints are invariant to outer-loop decisions, so the proposed architecture remains a receding-horizon MPC in the classical sense \cite{rawlings2017mpc,MAYNE2000789} with a supervisory tuning layer above it.

Importantly, the adaptive supervisory layer does not directly command thrust actions. Instead, it modifies high-level optimization parameters governing the MPC problem, including tracking weights, control penalties, prediction horizons, and safety-related objective terms. The constrained receding-horizon optimization structure itself remains unchanged throughout execution.

This separation between adaptive strategic tuning and low-level constrained control preserves the interpretability and constraint-awareness of the MPC formulation while enabling adaptive behavior in dynamically evolving orbital interactions.

The proposed framework was evaluated within the AIAA Capture the Satellite competition environment based on the KSPDG framework and the associated SpaceGym environment \cite{Allen2023SpaceGym}. The considered scenario involved a competitive multi-agent RPO interaction between a pursuing spacecraft (\textit{Bandit}), a target vehicle (\textit{Lady}), and an adversarial spacecraft (\textit{Guard}), where the corresponding guidance and control strategy achieved first-place performance during the final competition phase. Experimental results demonstrated robust adaptive MPC behavior together with improved success-rate trends relative to representative fixed-parameter MPC baselines while maintaining real-time computational feasibility and safety-aware operation.

Although optimization parameters are updated adaptively during execution, the control inputs are always generated through the repeated online solution of a constrained finite-horizon MPC problem using the current system state, prediction model, and active constraint set. Consequently, the supervisory adaptation process operates at the tuning level rather than replacing the underlying MPC formulation.

The main contributions of this work are summarized as follows:

\begin{itemize}

\item An AI-augmented MPC framework for adversarial RPO that combines constrained receding-horizon optimization with adaptive supervisory parameter tuning while preserving the underlying MPC structure;

\item A supervisory adaptation strategy that operates on physically interpretable MPC parameters rather than directly generating control commands, enabling adaptive behavior while maintaining transparency and constraint-aware operation;

\item An interpretable autonomy framework that relates adaptive controller behavior to physically meaningful quantities, including safety penalties, minimum-separation objectives, and evolving orbital interaction geometry;

\item Large-scale Monte Carlo validation within the official KSPDG Capture-the-Satellite environment, demonstrating improved mission performance and adaptive adversarial maneuvering behavior relative to fixed-parameter MPC baselines.

\end{itemize}

\section{Related Work}

MPC has been extensively studied for spacecraft RPO due to its ability to explicitly handle actuator limitations, collision-avoidance constraints, and trajectory optimization within a unified receding-horizon framework \cite{pesce_modern_2022,starek2017fast}. Additional MPC-based proximity operations frameworks have investigated guaranteed feasibility and thruster-constrained maneuver planning for small-body and orbital rendezvous scenarios \cite{Carson2006MPC}. Linearized relative orbital dynamics based on CW formulations are particularly attractive for real-time guidance and optimization because they preserve computational tractability while providing sufficiently accurate local relative motion representations for many proximity operations scenarios \cite{curtis2019orbital}. Automated rendezvous and docking architectures have historically relied on constrained guidance and navigation formulations capable of preserving operational safety, fault tolerance, and real-time maneuvering feasibility during close-proximity operations \cite{Fehse_2003}.

Early applications of constrained optimization and MPC to spacecraft proximity operations demonstrated the ability to explicitly handle actuator limitations, safety constraints, and fuel-efficiency objectives within a unified guidance framework. For example, \citet{richards2002spacecraft} investigated constrained trajectory optimization for spacecraft formation flying under state and actuator limitations, while \citet{breger2006mpc} developed fuel-optimal rendezvous trajectories with passive collision-avoidance constraints for autonomous spacecraft proximity operations.

More recently, learning-based and data-driven methods have been explored for adaptive spacecraft guidance, autonomous rendezvous, and multi-agent maneuver planning in dynamically evolving orbital environments. Recent studies have investigated reinforcement-learning-enhanced MPC architectures and adaptive decision-making frameworks that combine optimization-based control with data-driven adaptation under uncertainty \cite{Federici2026RLEMPC,aerospace12090837}. \citet{DBLP:journals/corr/HeBKD16} demonstrated that reinforcement learning combined with opponent-modeling strategies can improve adaptive decision-making in competitive multi-agent environments, highlighting the potential of learning-based approaches for dynamically evolving adversarial interactions. However, end-to-end learned control architectures often suffer from limited interpretability and difficulty enforcing hard safety constraints, which remains a major concern for safety-critical orbital robotics applications. This has motivated growing interest in safe reinforcement learning and safe learning control frameworks that explicitly incorporate safety constraints, robustness, and certifiable decision-making capabilities into learning-based systems~\cite{JMLR:v16:garcia15a, DBLP:journals/corr/abs-2108-06266}.

To address some of the limitations of purely learning-based control architectures, several recent works have investigated combinations of optimization and learning in order to improve adaptability while preserving safety, structure, and constraint-awareness within autonomous control systems \cite{hewing2020learning,Koller2018LearningBasedMP}. In particular, learning-enhanced control frameworks have been explored to improve prediction quality, online parameter adaptation, and strategic decision-making behavior while retaining optimization-based control structures \cite{kiumarsi2018reinforcement}. Similarly, multi-agent trajectory optimization and differential game formulations have demonstrated the importance of explicitly modeling interactions among autonomous agents operating in dynamically coupled environments \cite{FridovichKeil2019EfficientIL}.

The framework proposed in this work differs from end-to-end learning approaches in that the adaptive layer does not directly generate control actions. Instead, adaptation is restricted to the modification of high-level optimization parameters governing the MPC problem. The constrained receding-horizon optimization process, orbital dynamics model, and safety constraints remain active throughout execution. This preserves the interpretability and constraint-aware structure of the controller while enabling adaptive strategic behavior in adversarial RPO scenarios.

\section{Problem Formulation}

The considered RPO scenario involves three spacecraft operating in close relative motion:

\begin{itemize}
    \item a pursuing spacecraft, denoted as the \textit{Bandit};
    \item a target spacecraft, denoted as the \textit{Lady};
    \item an adversarial spacecraft, denoted as the \textit{Guard}.
\end{itemize}

The objective of the \textit{Bandit} is to approach the \textit{Lady} while avoiding unsafe proximity with the \textit{Guard}. The scenario therefore combines target tracking, constrained maneuvering, and adversarial interaction within a single constrained autonomous guidance problem. In the official competition environment, the \textit{Lady} and \textit{Guard} spacecraft behaviors were governed by the challenge simulation framework and could evolve dynamically throughout the interaction. The resulting interaction scenario required the \textit{Bandit} spacecraft to approach the target while reacting to dynamically evolving adversarial behavior from the \textit{Guard} vehicle together with periodic evasive maneuvering behavior from the \textit{Lady} spacecraft.

The relative dynamics used by the controller are formulated using CW dynamics about a circular reference orbit \cite{clohessy1960terminal}. The reference orbit is not intended to imply that all vehicles remain fixed on identical circular trajectories; rather, it provides the local orbital frame and mean motion used to construct a computationally efficient linearized prediction model. The spacecraft states are represented in a Local Vertical Local Horizontal (LVLH) frame associated with this reference orbit, while the MPC tracking problem is expressed using relative coordinates with respect to the \textit{Lady} spacecraft.

At each control update, the \textit{Bandit} relative state used by the controller is defined as in Eq.~\ref{eq:state}.

\begin{equation}
\label{eq:state}
x(k) =
\bigl[
r_x \quad r_y \quad r_z \quad v_x \quad v_y \quad v_z
\bigr]^\tp.
\end{equation}

$r_x$, $r_y$, and $r_z$ denote the \textit{Bandit} position relative to the \textit{Lady}, and $v_x$, $v_y$, and $v_z$ denote the corresponding relative velocity components. In this local tracking frame, the \textit{Lady} is treated as the instantaneous tracking reference, while the \textit{Guard} state is represented relative to the same frame for predictive safety assessment.

The control input is defined as in Eq.~\ref{eq:control}.

\begin{equation}
\label{eq:control}
u(k) =
\bigl[
u_x \quad u_y \quad u_z
\bigr]^\tp.
\end{equation}

$u_x$, $u_y$, and $u_z$ correspond to commanded translational accelerations along the local frame axes.

The autonomous guidance objective is formulated as a constrained multi-objective control problem. The controller seeks to:

\begin{itemize}
    \item minimize the relative distance between the \textit{Bandit} and the \textit{Lady};
    \item maintain safe separation between the \textit{Bandit} and the \textit{Guard};
    \item limit excessive control effort;
    \item satisfy actuator and safety constraints.
\end{itemize}

Safety constraints include thrust magnitude limits and predictive KOZ constraints around the \textit{Guard} spacecraft. The KOZ constraints are evaluated over the prediction horizon ($H_p$) whenever the predicted \textit{Guard} proximity falls within a predefined safety threshold. Slack variables are introduced to soften selected safety constraints and preserve numerical feasibility in highly adversarial configurations \cite{Kerrigan2000SoftCA}. These slack variables are penalized in the MPC cost function so that constraint relaxation is discouraged unless required to avoid infeasibility.

The proposed framework operates within a receding-horizon control architecture. At each timestep, updated relative state information is used to solve a constrained optimization problem, and only the first control action is applied. The supervisory layer adjusts selected optimization-level MPC parameters according to the active interaction geometry and mission state, but it does not directly generate thrust commands. The underlying constrained MPC structure, orbital dynamics model, actuator limits, and safety constraints remain active throughout execution.

\section{Methods}
\label{sec:methodology}

The overall control architecture introduced in Fig.~\ref{fig:architectureMPC}
combines predictive orbital dynamics, constrained receding-horizon optimization,
safety-aware constraint handling, and adaptive supervisory parameter tuning
within a unified autonomous guidance framework.

\subsection{Relative Orbital Dynamics}
\label{sec:relative_dynamics}

Relative motion prediction within the MPC framework is based on the CW, linearized orbital dynamics model \cite{clohessy1960terminal,starek2017fast}. The CW formulation provides a computationally efficient linear time-invariant approximation of relative spacecraft motion about a circular reference orbit and is widely used in RPO due to its compatibility with real-time guidance and optimization frameworks.

The linearized translational relative dynamics in the LVLH frame are governed by the CW equations reported in Eq.~\ref{eq:cw_dynamics}. In the LVLH convention adopted in this work, the $x$ axis denotes the radial direction, the $y$ axis denotes the along-track direction, and the $z$ axis denotes the orbit-normal direction.

\begin{equation}
\begin{aligned}
\ddot{x} - 2n\dot{y} - 3n^2x &= u_x, \\
\ddot{y} + 2n\dot{x} &= u_y, \\
\ddot{z} + n^2z &= u_z.
\end{aligned}
\label{eq:cw_dynamics}
\end{equation}

$u_x$, $u_y$, and $u_z$ are the commanded translational accelerations, and $n=\sqrt{\mu / r_0^3}$ is the mean motion associated with a circular reference orbit of radius $r_0$, with $\mu$ the gravitational parameter of the primary body.

The continuous-time dynamics are rewritten in state-space form as shown in Eq.~\ref{eq:ss_continuous}.

\begin{equation}
\dot{\mathbf{x}}(t) = A_c\mathbf{x}(t) + B_c\mathbf{u}(t).
\label{eq:ss_continuous}
\end{equation}

The continuous-time state and control vectors associated with the CW prediction model are defined in Eq.~\ref{eq:state_control}.

\begin{equation}
\mathbf{x}(t) =
\bigl[
x \quad y \quad z \quad \dot{x} \quad \dot{y} \quad \dot{z}
\bigr]^\tp,
\qquad
\mathbf{u}(t) =
\bigl[
u_x \quad u_y \quad u_z
\bigr]^\tp.
\label{eq:state_control}
\end{equation}

For real-time MPC implementation, the dynamics are discretized using a fixed controller update timestep $\Delta t$. A zero-order hold on $\mathbf{u}$ is assumed between samples, and the discrete-time prediction model is obtained via the standard matrix-exponential identity in Eq.~\ref{eq:zoh}.

\begin{equation}
\begin{bmatrix}
A_d & B_d \\
0   & I
\end{bmatrix}
\;=\;
\exp\!\left(
\begin{bmatrix}
A_c & B_c \\
0   & 0
\end{bmatrix}
\Delta t
\right).
\label{eq:zoh}
\end{equation}

This yields the discrete-time prediction model in Eq.~\ref{eq:ss_discrete}.

\begin{equation}
\mathbf{x}_{k+1} = A_d \mathbf{x}_k + B_d \mathbf{u}_k.
\label{eq:ss_discrete}
\end{equation}

$A_d \in \mathbb{R}^{6 \times 6}$ and $B_d \in \mathbb{R}^{6 \times 3}$ are the discrete-time state transition and control matrices used by the predictive controller.

The CW model was selected because it preserves the convex structure required for efficient Quadratic Programming (QP) while remaining sufficiently accurate for the local relative motion conditions encountered during the considered RPO scenario. Although the operational environment in the competition framework may include nonlinear and dynamically evolving interactions among spacecraft, the linearized prediction model provided adequate real-time predictive capability while maintaining moderate computational cost.

\subsection{Model Predictive Control Formulation}
\label{sec:mpc_form}

At each control update, the proposed controller solves a constrained finite-horizon optimization problem using the discrete-time prediction model defined in Eq.~\ref{eq:ss_discrete}. The MPC framework operates in a receding-horizon fashion: after solving the optimization problem, only the first control action is applied before the optimization is repeated using updated state information.

The resulting constrained finite-horizon optimization problem is reported in Eq.~\ref{eq:mpc_optimization_problem}, subject to the system dynamics, actuator limitations, and predictive safety constraints. The decision variables consist of the control sequence over the control horizon ($H_u$) and, when predictive KOZ constraints are active, the associated safety slack variables.

\begin{equation}
\label{eq:mpc_optimization_problem}
\begin{aligned}
\min_{\mathbf{u},\,\mathbf{s}} \quad & J \qquad 
\mathrm{s.t.}
&
\begin{cases}
\mathbf{x}(k+i+1)=A_d\mathbf{x}(k+i)+B_d\mathbf{u}(k+i), \\
\left\|\mathbf{u}(k+i)\right\| \le u_{\max}, \\
\left\|\mathbf{u}(k+i)-\mathbf{u}(k+i-1)\right\| \le \Delta u_{\max}, \\
g_{\mathrm{KOZ}}(\mathbf{x}(k+i)) + s_i \ge 0, \\
s_i \ge 0 .
\end{cases}
\end{aligned}
\end{equation}

The MPC objective function is formulated as the finite-horizon quadratic cost reported in Eq.~\ref{eq:mpc_cost}. The first term penalizes the predicted relative tracking error between the \textit{Bandit} and \textit{Lady} spacecraft over the prediction horizon. The matrices $Q$ penalize the relative tracking errors, while $R$ penalizes control effort over the control horizon. The slack variables $s_i$ soften selected predictive KOZ constraints over the active safety horizon $K_{\mathrm{KOZ}}$ and are penalized by $\lambda_s$ to discourage constraint relaxation unless required for feasibility.

\begin{equation}
\begin{aligned}
J
={}&
\sum_{i=1}^{H_p-1}
\left\|
\mathbf{x}_{B}(k+i)-\mathbf{x}_{L}(k+i)
\right\|_{Q}^{2}
+
\sigma_{H_p}
\left\|
\mathbf{x}_{B}(k+H_p)-\mathbf{x}_{L}(k+H_p)
\right\|_{Q}^{2}
\\
&+
\sum_{i=0}^{H_u-1}
\left\|
\mathbf{u}(k+i)
\right\|_{R}^{2}
+
\sum_{i=1}^{K_{\mathrm{KOZ}}}
\lambda_s s_i^{2}.
\end{aligned}
\label{eq:mpc_cost}
\end{equation}

\begin{figure}[hbt!]
\centering
\includegraphics[width=\textwidth]{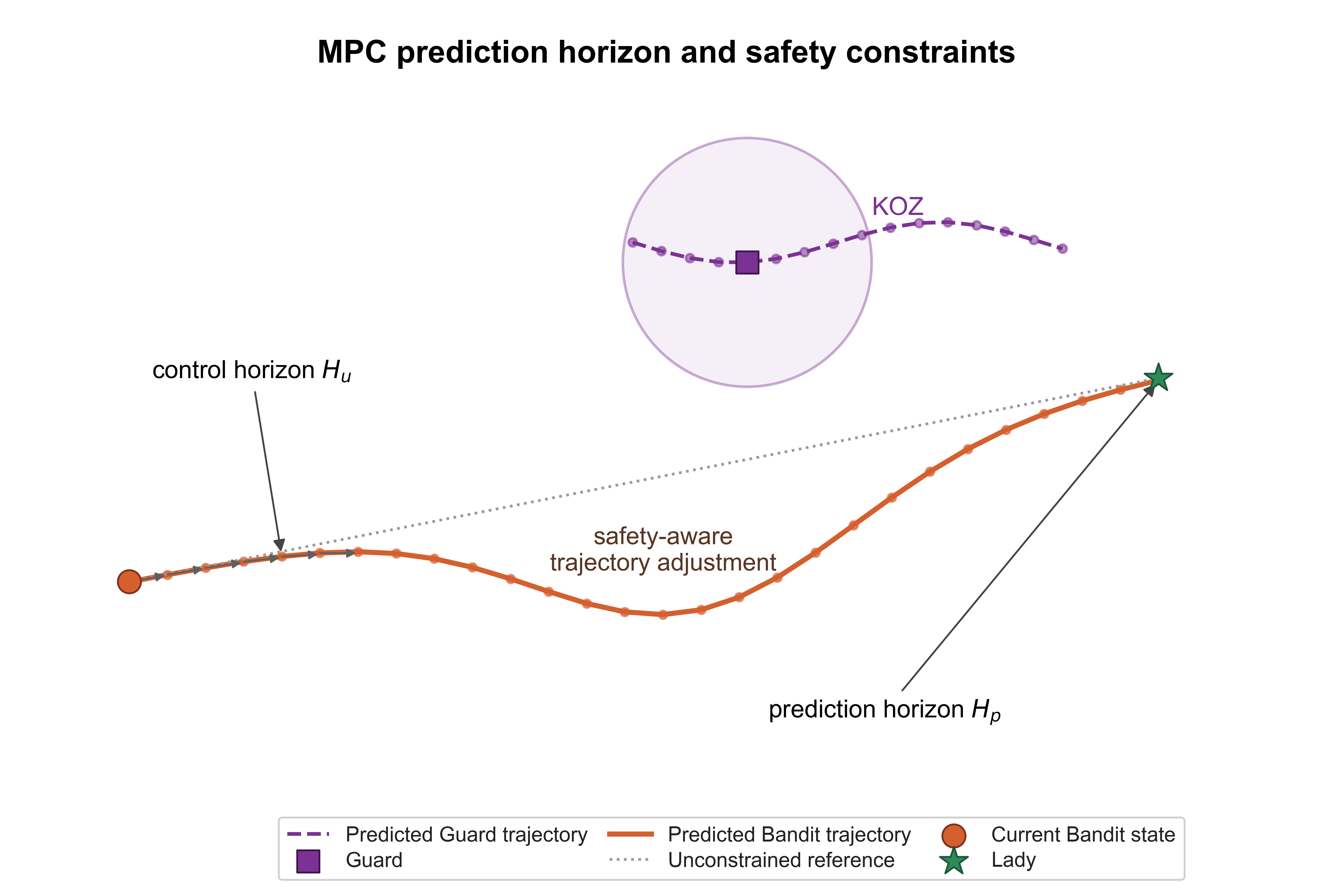}
\caption{Receding-horizon MPC prediction process and predictive safety-constraint evaluation.}
\label{fig:mpc_prediction_horizon}
\end{figure}

The receding-horizon prediction structure and the corresponding predictive safety constraint evaluation are illustrated in Fig.~\ref{fig:mpc_prediction_horizon}. The figure highlights how the optimizer evaluates future \textit{Bandit} and \textit{Guard} trajectories over the prediction horizon while enforcing safety-aware trajectory adjustments when KOZ interaction risk is present. At each control update, the optimizer propagates the predicted \textit{Bandit} trajectory over the prediction horizon $H_p$ using the discrete-time CW dynamics model. Predicted \textit{Guard} positions are simultaneously evaluated to identify potential KOZ violations. When predicted safety violations are detected within the horizon, safety constraints and associated slack variables are activated in the optimization problem, allowing the controller to trade rendezvous performance against safety preservation while maintaining feasibility.

\subsubsection{Stacked prediction}
$H_p$ denotes the prediction horizon and $H_u \le H_p$ the control horizon. 
The stacked predicted state and decision input sequences are defined in Eq.~\ref{eq:stacked_vectors}.

\begin{equation}
\mathbf{X}_k =
\begin{bmatrix}
\mathbf{x}_{k+1} \\ \vdots \\ \mathbf{x}_{k+H_p}
\end{bmatrix}
\in \mathbb{R}^{6 H_p},
\qquad
\mathbf{U}_k =
\begin{bmatrix}
\mathbf{u}_{k} \\ \vdots \\ \mathbf{u}_{k+H_u-1}
\end{bmatrix}
\in \mathbb{R}^{3 H_u}.
\label{eq:stacked_vectors}
\end{equation}

Repeated application of Eq.~\ref{eq:ss_discrete} yields the stacked prediction equation shown in Eq.~\ref{eq:stacked_pred}, where $\Phi \in \mathbb{R}^{6 H_p \times 6}$ and $\Gamma \in \mathbb{R}^{6 H_p \times 3 H_p}$ are the standard block-Toeplitz prediction matrices defined in Eq.~\ref{eq:Phi_Gamma}, and $T \in \mathbb{R}^{3 H_p \times 3 H_u}$ is a block lower-triangular lifting matrix mapping the $H_u$ decision controls to the $H_p$ applied controls, holding the input at the last decision value for $i \ge H_u$.

\begin{equation}
\mathbf{X}_k \;=\; \Phi\, \mathbf{x}_k \;+\; \Gamma\, T\, \mathbf{U}_k .
\label{eq:stacked_pred}
\end{equation}

\begin{equation}
\Phi =
\begin{bmatrix} A_d \\ A_d^2 \\ \vdots \\ A_d^{H_p} \end{bmatrix},
\qquad
\Gamma =
\begin{bmatrix}
B_d            & 0                & \cdots & 0 \\
A_d B_d        & B_d              & \cdots & 0 \\
\vdots         & \vdots           & \ddots & \vdots \\
A_d^{H_p-1} B_d & A_d^{H_p-2} B_d & \cdots & B_d
\end{bmatrix}.
\label{eq:Phi_Gamma}
\end{equation}

\subsubsection{Cost-function notation}

Using $\lVert v \rVert_M^2 \triangleq v^\top M v$, the weighting matrices in Eq.~\ref{eq:mpc_cost} are block-diagonal, as in Eq.~\ref{eq:QR}, and a terminal-cost multiplier $\sigma_{H_p} \ge 1$ is applied to the last prediction block to encourage longer-horizon strategic maneuver planning. The terminal multiplier takes the value $\mathrm{term}_{\mathrm{final}}$ when the \textit{Bandit} is inside the final-approach distance and $\mathrm{term}_{\mathrm{base}}$ otherwise.

\begin{equation}
Q = \operatorname{blkdiag}( w_{L,\mathrm{pos}}\, I_3,\; w_{L,\mathrm{vel}}\, I_3 ),
\qquad
R = w_U\, I_3.
\label{eq:QR}
\end{equation}

\subsubsection{Decision variable and parametric QP}

The decision variable is defined as the concatenation of the control sequence and the active slack vector as shown in Eq.~\ref{eq:decision_var}.

\begin{equation}
\mathbf{z} =
\begin{bmatrix} \mathbf{U}_k \\ \mathbf{s} \end{bmatrix}
\in \mathbb{R}^{3 H_u + K_{\mathrm{KOZ}}}.
\label{eq:decision_var}
\end{equation}

Substituting the stacked prediction model of Eq.~\ref{eq:stacked_pred} into the MPC cost function of Eq.~\ref{eq:mpc_cost} yields the convex parametric quadratic program given in Eq.~\ref{eq:parametric_qp}.

\begin{equation}
\mathbf{z}^{\ast}(\mathbf{x}_k, P_k)
\;=\;
\arg\min_{\mathbf{z}}
\frac{1}{2}\mathbf{z}^\tp H(P_k)\mathbf{z} + f(\mathbf{x}_k, P_k)^\tp \mathbf{z}
\;\;\text{s.t.}\;\;
A(\mathbf{x}_k, P_k)\,\mathbf{z} \le b(\mathbf{x}_k, P_k),\;
\mathbf{z}_{lb} \le \mathbf{z} \le \mathbf{z}_{ub}.
\label{eq:parametric_qp}
\end{equation}

$P_k \in \mathcal{P}$ is the supervisory parameter vector and enters only the QP data $(H,f,A,b)$, never replacing the optimization itself. With the notation introduced above and $\bar Q = I_{H_p} \otimes Q$ (with the last block scaled by $\sigma_{H_p}$) and $\bar R = w_U I_{3 H_u}$, the QP data take the closed forms reported in Eq.~\ref{eq:Hf}.

\begin{equation}
H \;=\;
\begin{bmatrix}
T^\tp \Gamma^\tp \bar Q\, \Gamma T + \bar R & 0 \\
0 & \lambda_s\, I_{K_{\mathrm{KOZ}}}
\end{bmatrix},
\qquad
f \;=\;
\begin{bmatrix}
- T^\tp \Gamma^\tp \bar Q\, (\mathbf{X}_L^{\mathrm{ref}} - \Phi\, \mathbf{x}_k) \\
\mathbf{0}
\end{bmatrix}.
\label{eq:Hf}
\end{equation}

The linear inequality constraints \(A\mathbf{z} \le b\) encode the predictive safety constraints introduced in Section~\ref{sec:soft_constraints}. Translational actuator limits are enforced through the per-axis box constraints given in Eq.~\ref{eq:control_constraint}, matching the per-axis thrust capability of the \textit{Bandit}.

\begin{equation}
-u_{\max,j} \;\le\; \mathbf{z}_j \;\le\; +u_{\max,j} ,
\qquad j = 1,\ldots,3H_u.
\label{eq:control_constraint}
\end{equation}

Slack variables are additionally constrained according to $0 \le s_i \le s_{\max}$.

\subsubsection{Receding-horizon control law}
The applied control input is obtained from the first control block of the optimizer solution, as in Eq.~\ref{eq:rh_law}.

\begin{equation}
\mathbf{u}_k \;=\; \mathbf{z}^{\ast}_{1:3}(\mathbf{x}_k, P_k).
\label{eq:rh_law}
\end{equation}

The optimization problem of Eq.~\ref{eq:parametric_qp} is subsequently re-solved at time step $k{+}1$ using the updated spacecraft state and supervisory parameter vector. This corresponds to the standard receding-horizon MPC construction \cite{rawlings2017mpc,MAYNE2000789} with online parameter dependence on \(P_k\) rather than fixed controller tuning. When the optimization problem of Eq.~\ref{eq:parametric_qp} returns an infeasible status, a distance-scheduled PD control law saturated according to the actuator bounds of Eq.~\ref{eq:control_constraint} is applied as a numerical fallback strategy. In the reported experiments, this fallback mechanism was not activated during nominal mission phases.

The resulting constrained optimization problem is formulated primarily as a quadratic program and solved repeatedly online throughout closed-loop operation. The use of linearized CW dynamics together with quadratic cost terms and linearized predictive safety constraints preserves computational tractability and enables real-time execution within the competition environment.

At each control update, the controller solves a constrained finite-horizon optimization problem using the current spacecraft state, the active prediction model, and the current constraint set. Although selected optimization parameters may be modified online by the adaptive supervisory layer, the underlying MPC problem formulation, orbital dynamics model, actuator constraints, and predictive safety constraints remain embedded within the optimization problem throughout execution. Consequently, the adaptive mechanism modifies the tuning of the predictive controller rather than replacing the underlying constrained MPC structure itself.

\subsection{Safety Constraints and Feasibility Handling}
\label{sec:soft_constraints}

Safety preservation was treated as a primary design requirement of the proposed architecture. Rather than allowing the adaptive supervisory layer to directly generate control commands, all maneuvering actions remained constrained by the MPC optimization problem, actuator limitations, and predictive safety constraints throughout execution.

Predictive KOZ constraints were introduced to reduce unsafe proximity interactions between the \textit{Bandit} and \textit{Guard} spacecraft, as conceptually illustrated in Fig.~\ref{fig:mpc_prediction_horizon}. Both the KOZ and the surrounding safety buffer were modeled as spherical proximity regions centered on the predicted \textit{Guard} position. The inner KOZ defined the minimum allowable separation distance, while the larger surrounding safety buffer defined the region within which predictive safety analysis became active.

The \textit{Guard} relative trajectory was propagated over the MPC prediction horizon using the same CW-based prediction structure employed for the controlled \textit{Bandit} dynamics. At each prediction step, the optimizer evaluated the predicted relative separation between the two vehicles in order to determine whether future safety violations could occur over the receding horizon.

Importantly, predictive KOZ constraints were not enforced uniformly during all phases of the maneuver. Instead, the controller activated predictive KOZ analysis only when the predicted \textit{Guard} trajectory entered a larger predefined safety buffer region surrounding the active keep-out zone. Consequently, the safety buffer acted as an early-warning region within which the controller began evaluating future KOZ penetration risk before direct violation of the minimum-separation boundary could occur.

When the predicted \textit{Guard} trajectory remained outside the safety buffer region, the controller operated using nominal tracking-oriented MPC behavior without activating additional predictive KOZ constraint handling. As the \textit{Guard} entered the safety buffer region, predictive safety constraints and associated slack-variable handling became active over the dynamically relevant subset of the prediction horizon.

The use of a predictive safety buffer reduced unnecessary conservatism during nominal rendezvous phases while enabling earlier evasive maneuver planning during rapidly evolving adversarial interactions. Consequently, the controller could maintain more aggressive target-approach behavior when the \textit{Guard} spacecraft remained sufficiently distant, while progressively prioritizing safety-aware maneuvering as predicted \textit{Guard} proximity increased.

Once the safety buffer becomes active, predictive KOZ constraints are enforced over the subset \(i = 1,\ldots,K_{\mathrm{KOZ}}\) of the prediction horizon associated with potentially unsafe future interaction intervals. \(\mathbf{r}_{BG}^{\mathrm{nom}}(k+i)\) denotes the predicted \textit{Bandit}-\textit{Guard} relative position at prediction step \(i\) under the current nominal open-loop trajectory. The corresponding line-of-sight unit vector \(\hat{\mathbf{n}}_i\) together with the nominal safety-margin quantity \(h_0^i\) are defined in Eq.~\ref{eq:los}.

\begin{equation}
\hat{\mathbf{n}}_i \;=\;
\frac{\mathbf{r}_{BG}^{\mathrm{nom}}(k+i)}{\bigl\lVert \mathbf{r}_{BG}^{\mathrm{nom}}(k+i) \bigr\rVert} ,
\qquad
h_0^i \;=\; \bigl\lVert \mathbf{r}_{BG}^{\mathrm{nom}}(k+i) \bigr\rVert - d_{\mathrm{KOZ}}.
\label{eq:los}
\end{equation}

A first-order expansion along \(\hat{\mathbf{n}}_i\) yields the linearized predictive safety condition reported in Eq.~\ref{eq:koz_constraint_lin}. In this formulation, \(M_i \in \mathbb{R}^{3 \times 3 H_u}\) maps the decision controls to the predicted \textit{Bandit} position deviation at prediction step \(i\), \(\gamma_K > 0\) is a decay coefficient regulating the rate at which the linearized constraint may approach the boundary, and \(s_i \ge 0\) is the corresponding slack variable.

\begin{equation}
- \hat{\mathbf{n}}_i^{\tp} M_i\, \mathbf{U}_k \;-\; s_i
\;\le\;
- \hat{\mathbf{n}}_i^{\tp}\, \mathbf{r}_{BG}^{\mathrm{nom}}(k+i)
\;-\; \gamma_K\, h_0^i.
\label{eq:koz_constraint_lin}
\end{equation}

The original nonlinear distance condition \(d_{BG}(k+i) + s_i \ge d_{\mathrm{KOZ}}\) is preserved logically, while the linearized form of Eq.~\ref{eq:koz_constraint_lin} preserves the convex structure of the underlying quadratic program required for real-time online solution. The slack variables are introduced only over the \(K_{\mathrm{KOZ}}\) prediction intervals with active predictive safety enforcement, thereby reducing optimization dimensionality while localizing constraint softening to dynamically relevant safety-critical interaction intervals.

The slack variables soften selected predictive safety constraints in situations where strict feasibility cannot be maintained due to highly adversarial geometric configurations, limited maneuver authority, or conflicting tracking and safety objectives. These slack variables are directly penalized through the MPC cost function, ensuring that constraint relaxation is minimized whenever feasible solutions exist.

Importantly, the slack variables do not remove the safety constraints from the optimization problem. Instead, they provide controlled and explicitly penalized feasibility relaxation in order to avoid optimizer infeasibility, unstable control behavior, or excessively aggressive evasive maneuvers during adversarial multi-agent interactions. This formulation allows the controller to preserve continuous closed-loop operation even when perfect simultaneous satisfaction of all tracking and safety objectives cannot be achieved.

Consequently, the adaptive supervisory layer was restricted to modifying high-level optimization parameters while the constrained MPC structure itself remained unchanged. The adaptive layer therefore could not directly bypass the orbital dynamics model, actuator limitations, or predictive safety constraints embedded within the optimization problem.

\subsection{Control Barrier Function safety filter}

As an optional inner-most safety mechanism, the MPC-computed control \(\mathbf{u}_{\mathrm{cmd}}\) can be passed through the discrete-time CBF filter defined in Eq.~\ref{eq:cbf_proj} before being applied to the actuator \cite{ames2017cbf}. The safety barrier function \(h(\mathbf{x})\), positive whenever the \textit{Bandit} spacecraft remains outside the KOZ, is defined as $h(\mathbf{x}) = d_{BG} - d_{\mathrm{KOZ}}$. Its time derivative along the relative dynamics is given in Eq.~\ref{eq:hdot}.

\begin{equation}
\dot h = \hat{\mathbf{n}}^{\tp} (\mathbf{v}_B - \mathbf{v}_G),
\qquad
\hat{\mathbf{n}} = (\mathbf{r}_B - \mathbf{r}_G)/\lVert \mathbf{r}_B - \mathbf{r}_G\rVert.
\label{eq:hdot}
\end{equation}

A discrete-time higher-order CBF condition with gains \(k_0,k_1 > 0\) is imposed according to Eq.~\ref{eq:cbf}, where \(\mathbf{a}_G^{\mathrm{now}}\) denotes the current PD-pursuit estimate of the adversary acceleration.

\begin{equation}
\hat{\mathbf{n}}^{\tp} \mathbf{u}_{\mathrm{safe}}
\;\ge\;
\beta(\mathbf{x}_k), \qquad
\beta(\mathbf{x}_k) \;\triangleq\;
- \hat{\mathbf{n}}^{\tp} \mathbf{a}_G^{\mathrm{now}} - k_1\, \dot h - k_0\, h.
\label{eq:cbf}
\end{equation}

The filtered control input is obtained from the projection quadratic program reported in Eq.~\ref{eq:cbf_proj}.

\begin{equation}
\mathbf{u}_{\mathrm{safe}} \;=\;
\arg\min_{\mathbf{u}\in\mathbb{R}^3}\;
\lVert \mathbf{u} - \mathbf{u}_{\mathrm{cmd}} \rVert^{2}
\;\;\text{s.t.}\;\;
\hat{\mathbf{n}}^{\tp} \mathbf{u} \ge \beta(\mathbf{x}_k),\;\;
-u_{\max} \le \mathbf{u} \le +u_{\max}.
\label{eq:cbf_proj}
\end{equation}

When the condition of Eq.~\ref{eq:cbf} is already satisfied by \(\mathbf{u}_{\mathrm{cmd}}\), the filter acts as the identity map. The primary safety guarantees in the implemented framework are provided by the constrained MPC formulation of Eq.~\ref{eq:parametric_qp} together with predictive safety evaluation and feasibility-preserving slack handling; the CBF filter is used as an additional defense layer.

Formal recursive feasibility and closed-loop stability guarantees under arbitrary adaptive parameter updates remain outside the scope of the present work. However, two design choices already provide partial guarantees. First, the parameter set \(\mathcal{P}\) over which the supervisory layer adapts is bounded by construction (Section~\ref{sec:adaptive_supervisory_layer}), so the inner-loop QP data of Eq.~\ref{eq:parametric_qp} depend continuously on a compact set of parameters at every step. Second, the soft-constraint formulation with bounded non-negative slacks introduced in Eq.~\ref{eq:decision_var} helps preserve optimization feasibility under strongly adversarial interaction conditions \cite{Kerrigan2000SoftCA}. Beyond these structural guarantees, robustness is validated empirically through the Monte Carlo campaign of Section~\ref{sec:mc}. The proposed framework therefore follows an engineering-oriented robustness validation approach rather than a theorem-driven stability analysis.

\subsection{Adaptive Parameter Layer}
\label{sec:adaptive_supervisory_layer}
The proposed framework incorporates an outer-loop adaptive layer designed to modify selected MPC optimization parameters online during closed-loop operation. Unlike end-to-end learned control architectures, the adaptive layer does not directly generate thrust commands or bypass the constrained optimization process. Instead, adaptation is restricted to high-level parameters governing the MPC objective and predictive behavior.

A representative subset of MPC parameters considered in the supervisory tuning framework is summarized in Table~\ref{tab:adaptive_parameters}. The implementation includes additional bounded parameters associated with KOZ horizon selection, control effort weighting, final-approach switching, guard-response shaping, and safety-constraint decay. All adapted parameters live in a bounded set $\mathcal{P}$ specified a priori, which ensures that the inner-loop QP remains well-posed under every supervisory decision.

\begin{table}[hbt!]
\caption{Representative subset of MPC parameters considered within the supervisory tuning framework.}
\label{tab:adaptive_parameters}
\centering
\renewcommand{\arraystretch}{1.3}
\begin{tabular}{p{0.24\linewidth} p{0.30\linewidth} p{0.36\linewidth}}
\hline
\textbf{Parameter Category} & \textbf{Adaptive Parameters} & \textbf{Functional Role} \\
\hline

Horizon handling &
$H_p$, $H_u$ &
Adjust the prediction and control horizons used by the receding-horizon optimizer. \\
\hline

Tracking objective &
$w_{L,\mathrm{pos}}$, $w_{L,\mathrm{vel}}$ &
Set the relative position and velocity tracking priorities with respect to the \textit{Lady} spacecraft. \\
\hline

Safety and KOZ handling &
$w_{\mathrm{KOZ}}$, $d_{\min}$ &
Tune the penalty and minimum-separation behavior associated with \textit{Guard} avoidance. \\
\hline

Terminal weighting &
$\mathrm{term}_{\mathrm{base}}$, $\mathrm{term}_{\mathrm{final}}$ &
Adjust terminal weighting during nominal and final approach phases. \\
\hline

Barrier filtering &
$k_0$, $k_1$ &
Shape the optional CBF-style distance and closing-rate safety response. \\
\hline

\end{tabular}
\end{table}

The resulting parameter updates influence the strategic behavior of the controller while preserving the underlying constrained MPC structure described in Sections~\ref{sec:mpc_form} and~\ref{sec:soft_constraints}.

\subsubsection{Offline stage}

A base parameter vector $P^{\ast} \in \mathcal{P}$ is obtained by minimizing the expected closed-loop episode return according to Eq.~\ref{eq:offline_obj}.

\begin{equation}
P^{\ast} \;\in\; \arg\min_{P \in \mathcal{P}}\;
\mathbb{E}_{\xi}\!\left[\, J\!\left(\pi_{\mathrm{MPC}}(\cdot; P),\; \xi\right) \,\right].
\label{eq:offline_obj}
\end{equation}

In Eq.~\ref{eq:offline_obj}, $\xi$ denotes an initial-condition seed drawn from the KSPDG environment distribution, $\pi_{\mathrm{MPC}}(\cdot;P)$ is the receding-horizon control law of Eq.~\ref{eq:rh_law} instantiated with parameter $P$, and $J$ is the mission cost returned by the environment (lower is better). The expectation of Eq.~\ref{eq:offline_obj} is non-convex in $P$ and not analytically differentiable; it is therefore approximated by sample average over $n_{ep}$ episodes and optimized using derivative-free search methods. Two optimization approaches were implemented: the Cross-Entropy Method (CEM) \cite{deboer2005cem,DBLP:journals/corr/abs-2009-09043} and Covariance Matrix Adaptation Evolution Strategy (CMA-ES) \cite{hansen2001cmaes}. CEM is a population-based stochastic optimization method that iteratively samples candidate parameter vectors from a probability distribution, evaluates their closed-loop performance, selects an elite subset of the best-performing candidates, and updates the sampling distribution toward the statistics of this elite set. In this work, CEM was used to search over the bounded parameter set $\mathcal{P}$ and identify a baseline MPC parameter vector with favorable empirical closed-loop performance; interested readers are referred to the original CEM literature for further details \cite{deboer2005cem}. For the results reported in this work, the offline parameter vector $P^\ast$ was obtained using CEM, while CMA-ES was retained as an implemented alternative optimization strategy.

\subsubsection{Online stage}

During deployment, the controller operates using the offline-optimized parameter set \(P^\ast\) together with a set of geometry-driven supervisory adaptation mechanisms. These mechanisms modify selected MPC optimization parameters according to the evolving interaction geometry while preserving the underlying constrained MPC formulation, orbital dynamics model, actuator constraints, and predictive safety constraints.

The implemented online adaptation logic consists of three geometry-dependent supervisory mechanisms:

\begin{itemize}

\item \textbf{Proximity-based minimum-separation inflation.}
The effective minimum-separation objective \(d_{\min}\) is inflated only when the current predicted \textit{Bandit}--\textit{Guard} separation \(d_{BG}\) falls below \(1.25\,d_{\min}\). A normalized proximity factor is first computed according to Eq.~\ref{eq:prox_factor}.

\begin{equation}
\mathrm{prox}=
\mathrm{clip}\!\left(
\frac{1.25\,d_{\min}-d_{BG}}
     {1.25\,d_{\min}},
0,1
\right).
\label{eq:prox_factor}
\end{equation}

The effective minimum-separation objective is then updated according to Eq.~\ref{eq:dmin_inflation}.

\begin{equation}
d_{\min,\mathrm{eff}}
=
d_{\min}
\left(
1+0.08\,\mathrm{prox}
\right).
\label{eq:dmin_inflation}
\end{equation}

This mechanism increases the desired safety margin only when the \textit{Guard} spacecraft is meaningfully close to the \textit{Bandit}.

\item \textbf{Phase-dependent tracking-weight scaling.}
The relative-position and relative-velocity tracking weights are scaled according to the current \textit{Bandit}--\textit{Lady} separation \(d_{BL}\). Specifically, multipliers of \((4,25)\), \((8,50)\), and \((12,80)\) are applied to \((w_{L,\mathrm{pos}}, w_{L,\mathrm{vel}})\) when \(d_{BL}\) falls below 500 m, 200 m, and 80 m, respectively. An additional final-approach condition applies intermediate multipliers of \((2.5,12)\). The resulting effective weights enter the block-diagonal MPC weighting matrix \(Q\) used in the quadratic cost.

\item \textbf{Conditional KOZ-weight modulation.}
The KOZ safety weight \(w_{\mathrm{KOZ}}\) is adjusted during QP construction and relaxation according to the active interaction geometry and solver feasibility. If the capture lane is open, the KOZ horizon is disabled and the effective KOZ weight is set to zero. If the nominal QP is infeasible, a first relaxation level reduces the effective KOZ weight to \(0.5\,w_{\mathrm{KOZ}}\); if infeasibility persists, a second relaxation level reduces it to \(0.3\,w_{\mathrm{KOZ}}\). As a last-resort guidance mode, the KOZ terms are disabled by setting the effective KOZ weight to zero. This rule reduces excessive conservatism only when required to recover QP feasibility while preserving the nominal safety-aware formulation whenever feasible.

\end{itemize}

These supervisory mechanisms operate directly on interpretable MPC optimization parameters and are evaluated throughout closed-loop execution. Consequently, adaptive behavior emerges through geometry-dependent parameter modulation rather than through direct modification of the underlying control law.

\subsection{Real-Time Control Architecture}
In deployment, the controller operates according to the closed-loop architecture illustrated in Fig.~\ref{fig:architectureMPC}.

\begin{enumerate}
    \item Acquisition of updated relative spacecraft state information;
    
    \item Propagation of the MPC prediction model over the receding horizon;
    
    \item Evaluation of predictive safety constraints and keep-out zone conditions;
    
    \item Online solution of the constrained parametric QP in Eq.~\ref{eq:parametric_qp};
    
    \item Optional CBF projection of the first optimized control action via Eq.~\ref{eq:cbf_proj};
    
    \item Application of the projected control to the \textit{Bandit} spacecraft;
    
    \item Geometry-driven supervisory update of the active MPC parameters according to the mechanisms described in Section~\ref{sec:adaptive_supervisory_layer}.
\end{enumerate}

This closed-loop process was repeated continuously throughout the maneuvering scenario. Prior to deployment, the baseline parameter vector \(P^\ast\) was obtained through offline optimization across multiple simulation episodes generated under varying initial orbital configurations and adversarial encounter geometries, as described in Section~\ref{sec:adaptive_supervisory_layer}. This process exposed the optimization procedure to a diverse set of interaction conditions and mission outcomes, allowing the selection of a parameter set with robust closed-loop performance across the considered scenario distribution. During execution, the controller operates using the offline-optimized parameter set together with the geometry-driven supervisory adaptation mechanisms described previously. The online supervisory layer modifies selected MPC tuning parameters according to the current interaction geometry while preserving the underlying constrained MPC formulation.

The computational complexity of the optimization problem remained compatible with real-time execution due to the use of linearized CW dynamics, quadratic objective functions, and moderate prediction horizon sizes. The constrained MPC formulation, together with the safety handling mechanisms described in Section~\ref{sec:soft_constraints}, remained active throughout all controller updates independently of the adaptive parameter modifications. The implemented architecture preserves a layered autonomy structure in which geometry-driven supervisory adaptation, constrained optimization, orbital dynamics propagation, and safety evaluation remain functionally separated. This separation improves interpretability while maintaining the constraint-aware behavior required for safety-critical orbital operations.

\section{Experimental Setup}

\subsection{KSPDG Capture-the-Satellite Scenario}

KSPDG is the official environment for the 2026 AIAA Capture the Satellite challenge \cite{Allen2023SpaceGym}. The considered scenario consists of an adversarial \textit{Lady}--\textit{Bandit}--\textit{Guard} RPO configuration involving three spacecraft interacting simultaneously within a closed-loop orbital environment. The controlled vehicle, denoted as the \textit{Bandit}, represented the autonomous pursuing spacecraft implementing the proposed adaptive MPC architecture. The \textit{Lady} spacecraft acted as the rendezvous target, while the \textit{Guard} spacecraft represented an adversarial vehicle attempting to intercept or obstruct the \textit{Bandit} during proximity operations with undisclosed control policy \cite{Allen2023SpaceGym}.

Each evaluation trial consisted of a complete KSPDG simulation run in which the \textit{Bandit} spacecraft attempted to safely approach the \textit{Lady} spacecraft while avoiding unsafe interactions with the \textit{Guard} vehicle using the proposed adaptive MPC framework described in Section~\ref{sec:methodology}.

The orbital guidance problem was formulated entirely in relative coordinates within the LVLH frame associated with the target spacecraft. At each control update, the controller received updated relative state information from the simulation environment and solved the corresponding constrained finite-horizon MPC problem online.

\subsection{Controller Configurations}

Three controller configurations were evaluated in order to compare the effect of adaptive supervisory parameter tuning on closed-loop rendezvous performance. Table~\ref{tab:controller_configs} summarizes the primary characteristics of the evaluated controller configurations used throughout the experimental campaign.

\begin{table}[hbt!]
\caption{Summary of evaluated controller configurations.}
\label{tab:controller_configs}
\centering
\renewcommand{\arraystretch}{1.3}
\begin{tabular}{p{0.20\linewidth} p{0.16\linewidth} p{0.20\linewidth} p{0.30\linewidth}}
\hline
\textbf{Configuration} & \textbf{Adaptive Tuning} & \textbf{Safety Handling} & \textbf{Control Characteristics} \\
\hline

Naive MPC &
No &
Minimal baseline safety handling. &
Basic constrained MPC with fixed nominal parameters. \\
\hline

Fixed Agent &
No &
Predictive KOZ constraints + slack handling. &
Manually tuned constrained MPC with fixed optimization parameters. \\
\hline

Adaptive Agent &
Yes &
Predictive KOZ constraints + slack handling. &
Adaptive constrained MPC with CEM-tuned base parameters and geometry-driven online supervisory modulation. \\
\hline

\end{tabular}
\end{table}

The first configuration consisted of a naive MPC controller operating with nominal fixed optimization parameters and without adaptive parameter updates. This configuration relied exclusively on the baseline MPC formulation throughout execution and provided a reference case for evaluating the effect of predictive safety handling and adaptive supervisory parameter tuning.

The second configuration employed a fixed-parameter constrained MPC controller incorporating predictive KOZ constraints, slack-variable feasibility handling, and manually tuned optimization parameters. The controller therefore combined constrained predictive safety handling with fixed supervisory tuning, while all optimization parameters remained constant throughout execution.

The third configuration employed the proposed adaptive MPC framework in which selected optimization-level controller parameters were updated online according to the evolving interaction conditions and supervisory adaptation logic described in Section~\ref{sec:adaptive_supervisory_layer}. The adaptive layer modified parameters including tracking weights, safety penalties, prediction horizon settings, and minimum KOZ separation objectives while preserving the underlying constrained MPC structure.

All controller configurations operated using the discrete-time CW prediction model introduced in Section~\ref{sec:relative_dynamics} together with repeated online quadratic program optimization over a receding prediction horizon.

Parameter adaptation was performed through a combination of offline derivative-free tuning (Eq.~\ref{eq:offline_obj}) and online geometry-driven supervisory parameter modulation based on the evolving orbital interaction conditions.

\subsection{Evaluation Protocol}
\label{sec:mc}

A total of 1000 closed-loop simulation runs were executed, within the KSPDG environment, using varying initial interaction conditions and adversarial encounter geometries. Aggregate Monte Carlo outcome statistics and representative trajectory comparisons are reported in Section~\ref{sec:results}. In addition, the controller-configuration comparison was performed over 25 runs per controller configuration, while the large-scale robustness analysis focused on the 1000-run adaptive MPC Monte Carlo campaign. During each simulation run, the \textit{Bandit} spacecraft attempted to safely rendezvous with the \textit{Lady} spacecraft while avoiding interception or unsafe proximity interactions with the \textit{Guard} vehicle.

Table~\ref{tab:mc_metrics} summarizes the primary evaluation objectives together with the corresponding Monte Carlo performance metrics used throughout the experimental campaign.

\begin{table}[hbt!]
\centering
\caption{Monte Carlo evaluation objectives and representative performance metrics.}
\label{tab:mc_metrics}
\begin{tabular}{ll}
\hline
\textbf{Evaluation Objective} & \textbf{Representative Metric} \\
\hline
Rendezvous performance & Success rate, target-relative distance evolution \\
Safety preservation & Minimum \textit{Guard} separation distance, KOZ violations \\
Robustness under adversarial interactions & Failure rate across varying encounter geometries \\
Controller adaptability & Adaptive supervisory parameter evolution \\
Closed-loop feasibility & Feasible completion of closed-loop simulation runs \\
Numerical stability & Stable closed-loop trajectory behavior \\
\hline
\end{tabular}
\end{table}

Performance evaluation included both mission-level outcome metrics and trajectory-level safety analysis. The resulting dataset provided a large-scale empirical assessment of the proposed framework under realistic orbital interaction conditions. 
The Monte Carlo evaluation campaign was performed using controller configurations obtained after the offline parameter-optimization phase described in Section~\ref{sec:adaptive_supervisory_layer}. During Monte Carlo evaluation, the controller operated in deployment mode and no additional offline parameter training was performed across simulation runs.

Importantly, the Monte Carlo campaign was not intended to provide formal proof of recursive feasibility or closed-loop stability under arbitrary adaptive parameter updates. Instead, the evaluation was designed to provide engineering-oriented robustness validation through repeated closed-loop execution across a wide range of interaction scenarios.

\subsection{Logged Quantities and Interpretability Analysis}

During each closed-loop simulation run, the framework recorded spacecraft states, control actions, MPC optimization parameters, predictive safety constraint activity, KOZ interactions, QP relaxation events, and mission outcome statistics for subsequent robustness and interpretability analysis.

Because the adaptive supervisory layer operated on interpretable MPC tuning parameters rather than directly on low-level control actions, the resulting parameter trajectories could be directly correlated with evolving orbital geometry, adversarial interactions, and safety conditions throughout execution.

The resulting dataset enabled post-run analysis of adaptive parameter evolution, safety-performance trade-offs, constraint activation behavior, and closed-loop robustness characteristics under dynamically evolving adversarial interaction conditions.

Algorithm~\ref{alg:interpretability_pipeline} summarizes the interpretability-oriented logging and post-processing workflow used throughout the evaluation campaign. The recorded quantities include relative spacecraft geometry, control actions, supervisory MPC parameters, predictive safety activity, and mission outcomes, enabling post-run analysis of the relationship between interaction geometry, adaptive parameter evolution, and closed-loop controller behavior.

\begin{algorithm}[hbt!]
\caption{Interpretability-Oriented Logging and Analysis Procedure.}
\label{alg:interpretability_pipeline}
\begin{algorithmic}[1]

\For{each Monte Carlo simulation run}

\State Record relative spacecraft states
\(
\mathbf r_{BG},
\mathbf v_{BG},
\mathbf r_{BL},
\mathbf v_{BL}
\)
and control actions
\(
\mathbf u_k
\).

\State Record MPC optimization parameters
\(
w_{\mathrm{KOZ}},
d_{\min},
w_{L,\mathrm{pos}},
w_{L,\mathrm{vel}}
\)
and supervisory parameter updates.

\State Record predictive safety-constraint activity and KOZ interactions.

\State Record slack-variable activations and mission outcome statistics.

\EndFor

\State Analyze adaptive supervisory parameter-evolution trends.

\State Analyze controller behavior under varying interaction conditions.

\State Evaluate closed-loop robustness and safety behavior.

\end{algorithmic}
\end{algorithm}

\section{Results and Discussion}
\label{sec:results}

\subsection{Closed-Loop Behavior and Monte Carlo Outcome Characteristics}

\begin{figure}[hbt!]
\centering
\includegraphics[width=\textwidth]{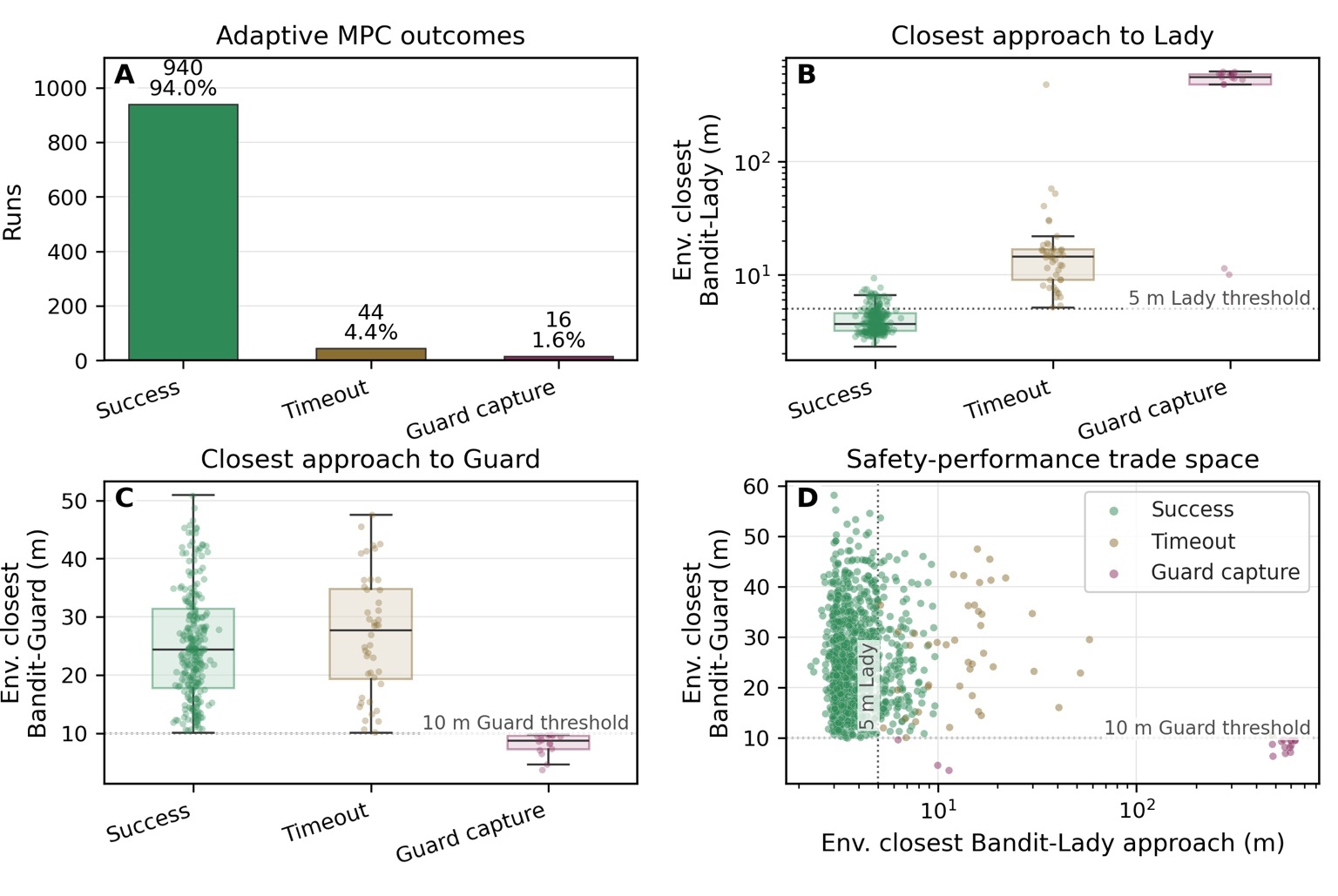}
\caption{Monte Carlo robustness evaluation for the 1000-run adaptive MPC campaign.}
\label{fig:mc}
\end{figure}

Monte Carlo outcome statistics and representative robustness characteristics generated by the proposed adaptive MPC framework are shown in Fig.~\ref{fig:mc}. Across the 1000-run evaluation campaign, the adaptive MPC controller achieved 940 successful runs, corresponding to a 94.0\% mission success rate, with 44 timeout cases and 16 \textit{Guard}-capture cases. The green dots represent success cases, the brown dots timeout cases, and the purple dots \textit{Guard}-capture cases. Panel (A) shows a success rate of 94.0\%, a \textit{Guard}-capture rate of 1.6\%, and a timeout rate of 4.4\%. Panels (B) and (C) show the closest \textit{Bandit}--\textit{Lady} and \textit{Bandit}--\textit{Guard} approach distances, respectively.

The outcome distributions and closest-approach statistics shown in Fig.~\ref{fig:mc}(B)--(D) indicate substantial variability in the relationship between rendezvous performance and adversarial separation behavior. In Fig.~\ref{fig:mc}(B), successful rendezvous cases achieve closest-approach distances below the 5 m target threshold, whereas timeout and \textit{Guard}-capture cases remain farther from the \textit{Lady}. In Fig.~\ref{fig:mc}(C), successful and timeout cases typically maintain closest \textit{Guard} separations between approximately 10 and 50 m, while \textit{Guard}-capture outcomes correspond to separations below the 10 m capture threshold. Successful runs consistently achieved close \textit{Bandit}-\textit{Lady} proximity, while exhibiting a broad range of closest \textit{Guard} separation distances ranging from near-threshold avoidance behavior to substantially larger safety margins.

Panel (D) shows the combined distribution of the two closest-approach distances. \textit{Guard}-capture cases are concentrated in the region characterized by small \textit{Bandit}--\textit{Guard} separation distances, whereas successful and timeout cases occupy a broader range of \textit{Guard} separations.

As shown in Fig.~\ref{fig:mc}(D), most \textit{Guard}-capture failures were additionally associated with relatively large final \textit{Bandit}-\textit{Lady} distances, suggesting that highly adversarial interaction geometries could simultaneously degrade rendezvous progression and safety preservation. Timeout outcomes exhibited more dispersed behavior across the safety--performance trade-space illustrated in Fig.~\ref{fig:mc}(D), including cases that remained relatively far from both spacecraft as well as cases that approached both the \textit{Lady} and \textit{Guard} vehicles more closely during the interaction. Several timeout cases additionally exhibited partial rendezvous progression while preserving relatively safe \textit{Guard} separation distances, suggesting that the controller occasionally favored conservative safety-preserving maneuvering behavior over aggressive terminal interception attempts under unfavorable interaction geometries.

The adaptive supervisory layer modified the strategic behavior of the controller through optimization-level parameter adaptation while preserving the constrained receding-horizon MPC structure responsible for generating the control actions. Consequently, all maneuvering behavior remained consistent with the imposed actuator limits, predictive safety constraints, and orbital dynamics model.

The resulting trajectory evolution further indicates that the adaptive controller maintained relatively smooth maneuvering behavior despite rapidly evolving adversarial conditions. The imposed control penalties and predictive horizon structure reduced excessive oscillatory thrust behavior while preserving sufficient responsiveness for collision avoidance and target pursuit.

\begin{figure}[hbt!]
\centering
\includegraphics[width=\textwidth]{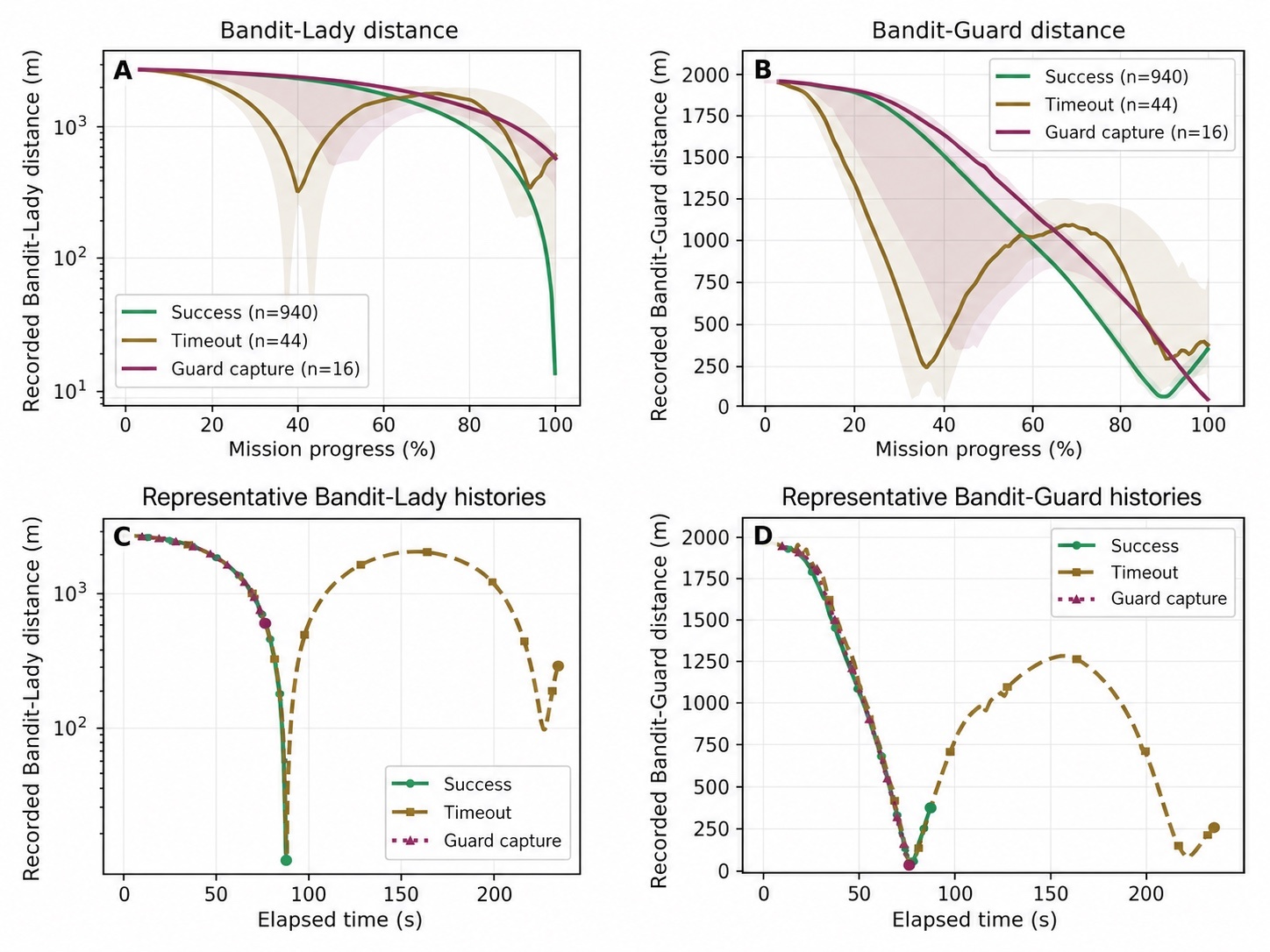}
\caption{Recorded \textit{Bandit}--\textit{Lady} and \textit{Bandit}--\textit{Guard} distance histories from the 1000-run adaptive MPC robustness dataset.}
\label{fig:distance_evolution}
\end{figure}

The mission-relative distance evolution observed during representative simulation runs is shown in Fig.~\ref{fig:distance_evolution}. The adaptive MPC framework progressively reduced the relative distance to the \textit{Lady} spacecraft while dynamically modifying the optimization priorities of the predictive controller according to the evolving adversarial interaction geometry.

The normalized mission-progress histories shown in Fig.~\ref{fig:distance_evolution}(A)--(B) correspond to population-level distance histories grouped by mission outcome and plotted over normalized mission progress for successful, timeout, and \textit{Guard}-capture trajectories. Successful runs typically exhibited monotonic reduction of \textit{Bandit}-\textit{Lady} distance while simultaneously delaying aggressive \textit{Guard} proximity until late mission phases. In contrast, timeout and \textit{Guard}-capture cases often displayed early safety-dominant interaction behavior followed by reduced rendezvous progression efficiency.

The representative histories shown in Panels (C) and (D) of Fig.~\ref{fig:distance_evolution} correspond to individual closed-loop simulation runs selected from the successful, timeout, and \textit{Guard}-capture outcome classes. These representative trajectories illustrate the different safety--performance trade-offs observed during the Monte Carlo campaign, including successful rendezvous progression, safety-preserving timeout behavior, and unsuccessful adversarial interactions leading to \textit{Guard} capture. Circular markers indicate the final recorded state of each trajectory. In several regions of the plots, overlapping trajectories partially obscure individual curves due to the large number of similar interaction histories contained within the dataset.

As illustrated by the representative single-run trajectories in Fig.~\ref{fig:distance_evolution}(C)--(D), temporary increases in target-relative distance were associated with elevated \textit{Guard} proximity and increased collision-risk conditions. Several timeout trajectories exhibited temporary increases in target-relative distance accompanied by recovery of \textit{Guard} separation margins, indicating that the predictive controller occasionally sacrificed short-term rendezvous efficiency in order to preserve longer-term safety feasibility. Representative successful trajectories shown in Panels (C) and (D) of Fig.~\ref{fig:distance_evolution} additionally showed that periods of decreasing \textit{Bandit}-\textit{Guard} separation were frequently followed by rapid recovery of adversarial distance margins during the terminal rendezvous phase. In contrast, several \textit{Guard}-capture cases exhibited sustained reduction of \textit{Bandit}-\textit{Guard} separation without comparable recovery behavior, indicating unsuccessful evasive response during highly adversarial interaction phases. During these phases, the supervisory adaptation layer increased safety-oriented optimization priorities, including keep-out zone penalties and minimum-separation objectives, resulting in more conservative maneuvering behavior intended to preserve safety margins.

As the \textit{Guard} threat level decreased and the interaction geometry became more favorable, the adaptive supervisory layer progressively reduced safety-dominant weighting behavior and restored more aggressive rendezvous-oriented optimization priorities. This interpretation is consistent with the supervisory MPC parameter trends shown in Fig.~\ref{fig:adaptive_parameters}, which provide additional evidence of the transition from safety-oriented to rendezvous-oriented behavior.

Overall, the recorded distance histories indicate that the adaptive MPC framework was capable of dynamically balancing rendezvous progression and adversarial avoidance behavior under rapidly evolving orbital interaction conditions. The resulting closed-loop trajectories exhibited coordinated transitions between target-pursuit and safety-preserving maneuvering regimes while maintaining stable and physically consistent behavior throughout the evaluated interaction scenarios.

\subsection{Controller Configuration Comparison}

\begin{figure}[hbt!]
\centering
\includegraphics[width=\textwidth]{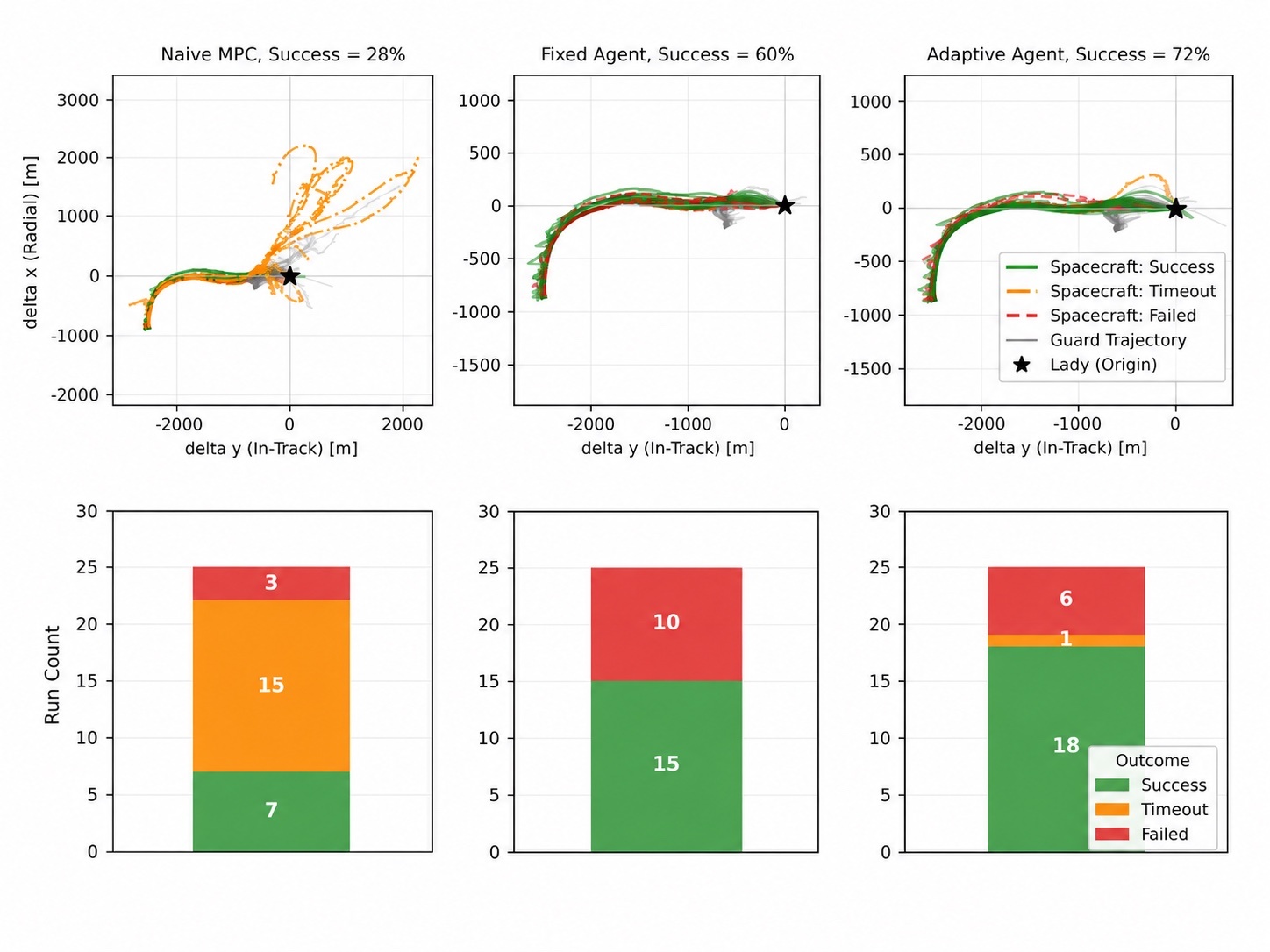}
\caption{Representative controller-configuration comparison across naive MPC, fixed-parameter MPC, and adaptive MPC cases.}
\label{fig:controller_comparison}
\end{figure}

Representative controller-comparison experiments were performed in order to evaluate the effect of supervisory adaptation on closed-loop rendezvous behavior under adversarial interaction conditions. Comparative results obtained over 25 simulation runs per controller configuration are shown in Fig.~\ref{fig:controller_comparison}. The figure illustrates qualitative differences in closed-loop trajectory behavior and observed mission outcomes under adversarial rendezvous conditions.

Figure~\ref{fig:controller_comparison} compares the three controller configurations over 25 simulation runs per configuration. The adaptive MPC controller achieved the highest success rate (72\%), followed by the fixed-parameter MPC controller (60\%) and the naive MPC controller (28\%). Correspondingly, the adaptive controller produced fewer unsuccessful mission outcomes and demonstrated more consistent rendezvous performance across the evaluated adversarial encounter conditions. These results illustrate the benefit of supervisory parameter adaptation under adversarial rendezvous conditions.

The adaptive supervisory layer enabled the controller to dynamically modify optimization priorities according to the evolving interaction geometry, producing visibly different maneuvering behavior relative to the fixed-parameter configurations during adversarial encounters. In contrast, fixed-parameter MPC configurations exhibited reduced adaptability during rapidly evolving adversarial encounters due to static weighting structures and fixed predictive behavior.

Across the evaluated comparison runs, the adaptive controller consistently maintained feasible closed-loop operation throughout dynamically evolving engagements. The imposed actuator limitations, predictive safety constraints, and feasibility-preserving constraint-relaxation handling remained active independently of the adaptive supervisory updates.

Differences between controller configurations were particularly visible during interaction phases involving aggressive \textit{Guard} interception trajectories. Under these conditions, the adaptive controller was able to adjust selected optimization parameters according to the evolving interaction geometry, enabling adaptive maneuvering behavior under rapidly changing adversarial conditions.

Importantly, the evaluation campaign was not intended to provide formal proof of recursive feasibility or closed-loop stability under arbitrary adaptive parameter adaptation. Instead, the reported results provide representative empirical evaluation through repeated closed-loop execution under dynamically evolving adversarial conditions.

No catastrophic controller divergence or persistent unstable oscillatory behavior was observed during the reported Monte Carlo campaigns. The resulting robustness characteristics therefore emerged from persistent constrained optimization, feasibility-preserving safety handling, and bounded supervisory adaptation rather than from theorem-driven stability guarantees.

\subsection{Adaptive Parameter Evolution and Interpretability}

\begin{figure}[hbt!]
\centering
\includegraphics[width=\textwidth]{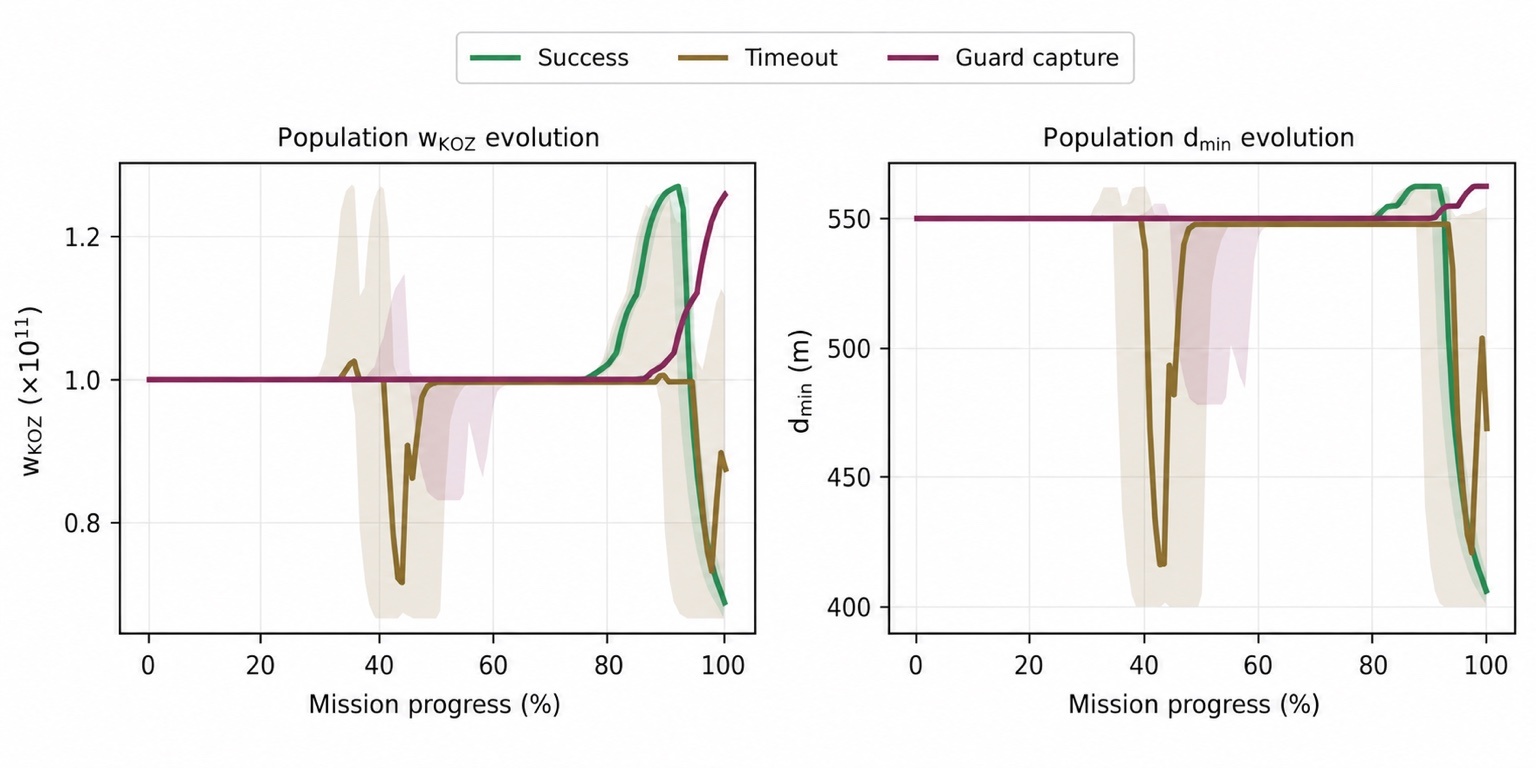}
\caption{Adaptive evolution of \(w_{\mathrm{KOZ}}\) and \(d_{\min}\) across successful, timeout, and \textit{Guard}-capture trajectories from the 1000-run adaptive MPC robustness dataset.}
\label{fig:adaptive_parameters}
\end{figure}

Representative adaptive supervisory parameter evolution trends are shown in Fig.~\ref{fig:adaptive_parameters}. Panel (A) reports the population-level evolution of the effective KOZ safety weight \(w_{\mathrm{KOZ}}\) as a function of normalized mission progress for successful, timeout, and \textit{Guard}-capture outcomes. Panel (B) shows the corresponding evolution of the effective minimum-separation objective \(d_{\min}\). Shaded regions indicate the 10th--90th percentile band computed across trajectories within each outcome class, while the solid curves represent the corresponding population-level trends. The figure illustrates how supervisory safety priorities evolve differently across successful and unsuccessful interaction scenarios. The adaptive MPC parameters summarized in Section~\ref{sec:adaptive_supervisory_layer} correspond to physically meaningful controller quantities, including the prediction horizon \(H_p\), the control horizon \(H_u\), the KOZ safety weight \(w_{\mathrm{KOZ}}\), and the minimum-separation objective \(d_{\min}\). Since the adaptive layer operates directly on interpretable MPC optimization parameters, the resulting controller behavior can be analyzed through the evolution of physically meaningful supervisory quantities including \(w_{\mathrm{KOZ}}\) and \(d_{\min}\).

The recorded parameter-evolution trends in Fig.~\ref{fig:adaptive_parameters} demonstrate how the supervisory adaptation layer dynamically modified safety-oriented optimization priorities according to the evolving adversarial interaction geometry. In the reported deployment results, the dominant online adaptive behavior was primarily associated with variation of the predictive KOZ safety weight \(w_{\mathrm{KOZ}}\) and the minimum-separation objective \(d_{\min}\), both of which directly influence the safety-performance balance of the constrained MPC problem.

There are distinct adaptive behaviors across successful, timeout, and \textit{Guard}-capture outcomes. In the successful outcome class (green curves), both \(w_{\mathrm{KOZ}}\) and \(d_{\min}\) frequently increased during late-mission interaction phases associated with reduced \textit{Bandit}-\textit{Guard} separation, consistent with the distance-history trends previously observed in Fig.~\ref{fig:distance_evolution}. As the \textit{Bandit} spacecraft approached the terminal rendezvous region near the \textit{Lady} spacecraft, the supervisory layer temporarily increased predictive safety priorities in order to preserve adversarial separation margins during the most collision-sensitive interaction phases. Following successful recovery of \textit{Guard} separation distance, both \(w_{\mathrm{KOZ}}\) and \(d_{\min}\) subsequently decreased, indicating a transition back toward more rendezvous-oriented optimization behavior once the immediate adversarial threat had been mitigated.

Timeout cases exhibited more prolonged conservative adaptation behavior, including sustained increases in minimum-separation objectives and delayed recovery toward more rendezvous-oriented optimization priorities. These trends are consistent with the previously observed distance-history evolution, where several timeout trajectories temporarily sacrificed rendezvous progression in order to preserve adversarial separation margins and maintain longer-term safety feasibility under unfavorable interaction geometries.

In contrast, several \textit{Guard}-capture trajectories exhibited comparatively delayed safety-parameter adaptation relative to the evolving adversarial geometry. In many of these cases, increases in \(w_{\mathrm{KOZ}}\) occurred only after substantial reduction of \textit{Bandit}-\textit{Guard} separation had already developed, suggesting that the timing of supervisory adaptation played an important role in determining whether adversarial avoidance behavior could be successfully recovered during late interaction phases.

Importantly, the adaptive layer modified interpretable optimization-level MPC quantities rather than directly generating low-level spacecraft thrust commands. Consequently, the observed adaptive behavior remained physically interpretable in terms of evolving predictive safety priorities, minimum-separation objectives, and rendezvous aggressiveness within the constrained MPC formulation.

\begin{figure}[hbt!]
\centering
\includegraphics[width=\textwidth]{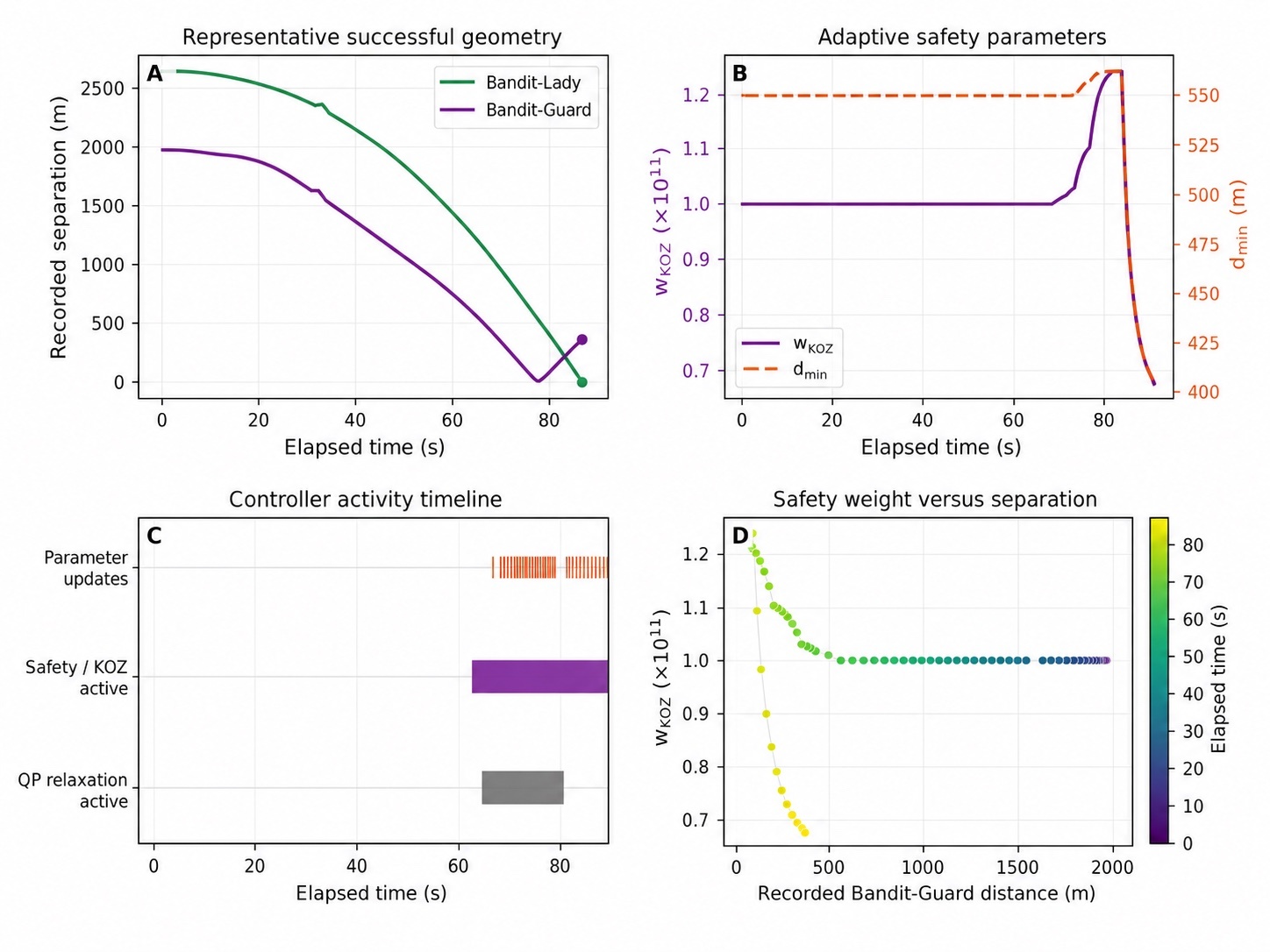}
\caption{Interpretability analysis of a representative successful adaptive MPC run showing geometry, supervisory adaptation, and controller behavior.}
\label{fig:parameter_geometry_interpretability}
\end{figure}

The recorded parameter trajectories further provide detailed post-run insight into the interaction between adversarial geometry evolution, supervisory adaptation, and constrained MPC behavior throughout closed-loop execution. Because the adaptive variables directly correspond to physically meaningful MPC optimization quantities, changes in controller behavior can be directly associated with evolving safety conditions and predictive maneuvering priorities. A representative geometry-parameter coupling analysis is shown in Fig.~\ref{fig:parameter_geometry_interpretability}, including the interaction geometry, adaptive safety parameters, controller activity, and the resulting safety-performance response.

Panel (A) of Fig.~\ref{fig:parameter_geometry_interpretability} shows the relative orbital interaction geometry during a representative successful rendezvous scenario. As the \textit{Guard} spacecraft progressively reduced separation distance relative to the \textit{Bandit} spacecraft prior to approximately \(t \approx 80\,\mathrm{s}\), the supervisory adaptation layer simultaneously increased both the predictive KOZ safety weight \(w_{\mathrm{KOZ}}\) and the minimum-separation objective \(d_{\min}\), as shown in Panel (B). These coordinated parameter increases indicate that the controller temporarily shifted the optimization problem toward more safety-dominant behavior during the most adversarial interaction phase.

As shown in Panel (B) of Fig.~\ref{fig:parameter_geometry_interpretability}, both \(w_{\mathrm{KOZ}}\) and \(d_{\min}\) decreased following recovery of \textit{Bandit}-\textit{Guard} separation distance after the closest-approach event, indicating that the supervisory adaptation layer progressively reduced safety-oriented optimization priorities once the immediate adversarial interaction became less critical. This transition was accompanied by simultaneous deactivation of QP relaxation activity together with coordinated reduction of both \(w_{\mathrm{KOZ}}\) and \(d_{\min}\), as shown in Panel (B) and Panel (C). The resulting behavior suggests that, following recovery of adversarial separation margins, the supervisory adaptation layer interpreted the restored interaction geometry as sufficiently safe to transition the optimization priorities back toward more aggressive rendezvous behavior with the \textit{Lady} spacecraft.

Panel (D) further illustrates the direct coupling between \(w_{\mathrm{KOZ}}\) and \textit{Bandit}-\textit{Guard} separation distance throughout the interaction. The resulting relationship demonstrates that the adaptive supervisory behavior remained strongly correlated with physically meaningful safety geometry rather than latent internal policy representations that may be more difficult to interpret directly in terms of physical safety objectives.
This constitutes an important distinction relative to fully end-to-end learned control architectures, where internal policy evolution may not directly correspond to interpretable control objectives or operational safety quantities \cite{DoshiVelez2017TowardsAR}. In the proposed framework, adaptive behavior can be directly traced to interpretable MPC optimization parameters associated with predictive safety preservation, adversarial avoidance behavior, and rendezvous aggressiveness, thereby enabling physically grounded post-run analysis of controller decision evolution throughout complex orbital interaction scenarios.

Across the evaluated Monte Carlo campaigns, the constrained MPC formulation remained active throughout execution despite the online supervisory adaptation process, allowing safety-aware predictive optimization behavior to be preserved under dynamically evolving adversarial interaction conditions.

\section{Conclusion}

This work presented an adaptive MPC framework for autonomous RPO in adversarial multi-agent orbital environments. The proposed architecture combined constrained receding-horizon optimization, linearized relative orbital dynamics, predictive safety handling, and supervisory adaptive parameter tuning within a unified closed-loop guidance framework.

Instead of directly generating spacecraft thrust commands as in some end-to-end learned control architectures, the proposed adaptive layer modifies interpretable optimization-level MPC parameters governing tracking performance, safety behavior, prediction-horizon selection, terminal weighting, and keep-out-zone handling within the constrained MPC formulation.
Consequently, the constrained MPC structure, orbital dynamics model, actuator limitations, and predictive safety constraints remained active throughout execution.

The proposed framework was evaluated within the official KSPDG Capture-the-Satellite competition environment through large-scale simulation campaigns involving dynamically evolving adversarial interactions between the \textit{Bandit}, \textit{Lady}, and \textit{Guard} spacecraft. The resulting closed-loop simulations demonstrated adaptive adversarial maneuvering behavior and favorable mission-outcome trends relative to representative fixed-parameter MPC configurations while preserving constraint-aware operation and real-time computational feasibility. Moreover, the presented adaptive MPC architecture was developed as part of the Embry-Riddle XDLab team submission, which achieved first place in the 2026 international AIAA Differential Game Capture the Satellite competition. 

The proposed framework provided a physically meaningful interpretation of the adaptive controller behavior.
The recorded supervisory parameter trajectories enabled physically grounded post-run analysis of controller behavior by directly relating adaptive optimization priorities to evolving orbital geometry and adversarial interaction conditions. The ability to directly relate adaptive supervisory behavior to physically meaningful MPC parameters improves transparency compared to fully end-to-end learned control policies and aligns with broader efforts toward interpretable and trustworthy autonomous systems for safety-critical applications.

Robustness was tested through persistent constrained optimization, feasibility-preserving safety handling, and extensive empirical Monte Carlo validation under adversarial interaction conditions. This engineering-oriented approach enabled the investigation of adaptive constrained autonomy while preserving the interpretability and modularity of the underlying MPC architecture. The present work does not provide formal recursive feasibility or closed-loop stability guarantees under arbitrary adaptive supervisory updates. 

More broadly, the proposed framework supports the perspective that adaptive supervisory parameter tuning can operate effectively on top of validated numerical control structures rather than directly replacing the underlying constrained guidance and control architecture.

Future work will investigate higher-fidelity nonlinear orbital dynamics models, additional adversarial interaction scenarios, onboard implementation considerations, and more advanced supervisory adaptation strategies together with stronger formal safety and stability guarantees for autonomous constrained guidance and control in complex multi-agent orbital environments.

\section*{Conflict of Interest}

The authors declare that the research was conducted in the absence of any commercial or financial relationships that could be construed as a potential conflict of interest.

\section*{Author Contributions}

LS proposed the adaptive MPC research concept, designed and implemented the control architecture, contributed to the adaptive supervisory training framework, conducted data processing and analysis activities, and wrote the manuscript. TB developed the adaptive supervisory training framework, conducted the simulation campaigns, and generated the manuscript figures. CK and DW supervised the research activities, contributed through technical guidance and scientific discussions, and provided feedback and revision support for the manuscript. All authors reviewed and approved the manuscript.

\section*{Data Availability Statement}

The datasets, simulation framework, and code developed for this study were generated within the research activities conducted at Embry-Riddle Aeronautical University. Access to portions of the simulation framework and associated implementation code may require institutional approval. Requests regarding data and research materials availability can be directed to the corresponding author.

\section*{Acknowledgments}

The authors would like to thank Ross E. Allen of MIT Lincoln Laboratory, the organizers of the AIAA Capture the Satellite challenge, and the Kerbal Space Program Differential Game development team for providing the competitive simulation environment used throughout this work and for technical discussions that helped motivate aspects of the present research. The authors also thank the reviewers and competition participants whose feedback and interactions contributed to the development and evaluation of the proposed framework. The authors acknowledge the use of ChatGPT (OpenAI) for limited language editing and manuscript polishing; all scientific content and final text were reviewed and approved by the authors.

\bibliography{ref}

\end{document}